\documentclass[runningheads]{llncs}
\usepackage{graphicx}

\usepackage{amsmath,amssymb} 
\usepackage{color}

\usepackage{gensymb}
\usepackage{mathtools}
\usepackage{comment}
\let\norm\undefined 
\DeclarePairedDelimiter\norm{\lVert}{\rVert}

\def\eg{\emph{e.g.}}
\def\ie{\emph{i.e.}}

\begin{document}

\title{Image-to-GPS Verification Through A Bottom-Up Pattern Matching Network} 
\titlerunning{Image-to-GPS Verification} 

\author{Jiaxin Cheng \and
Yue Wu \and
Wael Abd-Almageed \and
Prem Natarajan}
%

\authorrunning{Cheng Jiaxin et al.} 

\institute{University of Southern California, Information Sciences Institute \\
4746 Admiralty Way, Marina Del Rey, CA 90292, USA \\
\email{ \{chengjia,yue\_wu,wamageed,pnataraj\}@isi.edu }}


\maketitle

\begin{abstract}
The image-to-GPS verification problem asks whether a given image is taken at a claimed GPS location. In this paper, we treat it as an image verification problem -- whether a query image is taken at the same place as a reference image retrieved at the claimed GPS location. We make three major contributions: 1) we propose a novel custom bottom-up pattern matching (BUPM) deep neural network solution; 2) we demonstrate that the verification can be directly done by cross-checking a perspective-looking query image and a panorama reference image, and 3) we collect and clean a dataset of 30K pairs query and reference. Our experimental results show that the proposed BUPM solution outperforms the state-of-the-art solutions in terms of both verification and localization.

\end{abstract}

\section{Introduction}~\label{sec.introduction} 
In recent years we have seen many fake news stories, including but not limited to elections, natural disasters, protests, and riots. With the rapid growth of social networks and easy-to-use publishing applications on mobile devices, fake news can easily be produced and spread to social networks, and consequently to the entire world. Publishing fake news became a ``digital gold rush,''\footnote{\url{https://www.wired.com/2017/02/veles-macedonia-fake-news}} and detection tools need to be developed.

\begin{figure}[!h]
\centering\scriptsize
\setlength{\belowcaptionskip}{-20pt}
 \includegraphics[width=.95\linewidth]{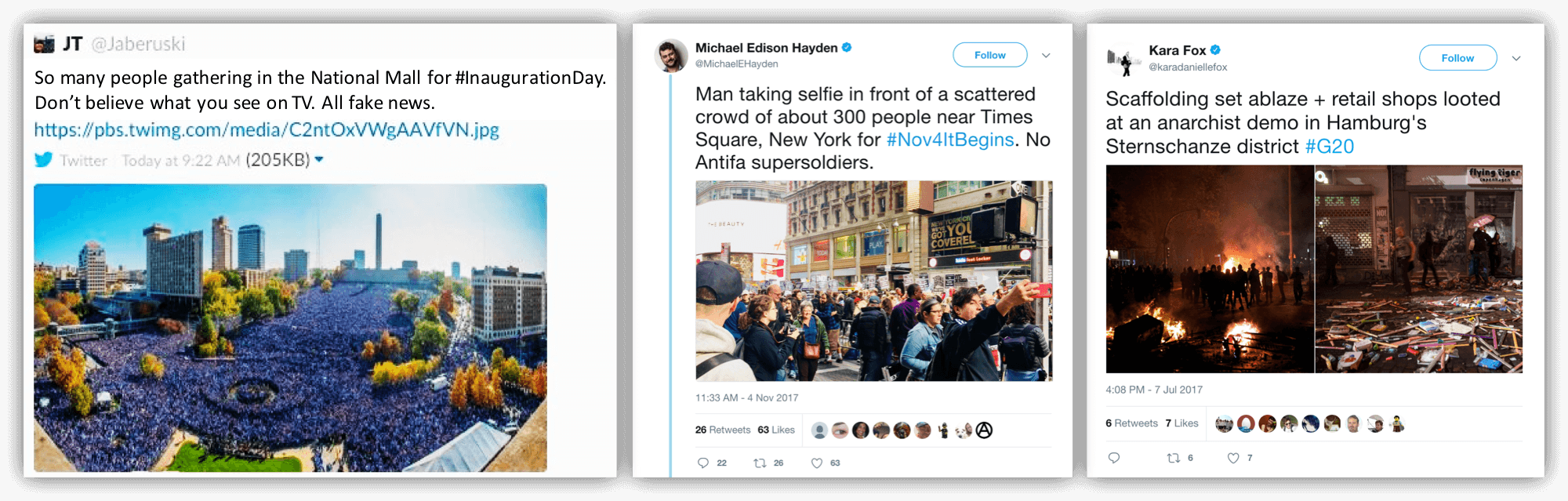}
\caption{Shall we trust these social network posts? Are these images taken at the claimed places?}\label{fig.fakenews}
\end{figure}

Many posts on social media are text-only, but it is common to see posts composed of both text and image/video (see samples in Fig.~\ref{fig.fakenews}), which is preferred by fake news posters, possibly because appealing photos makes fake news more convincing. However, this provides us extra opportunities to identify fake news, because one needs to tell more lies to make up one lie, but we only need to recognize one lie to conclude he/she is a lier. 

In this paper, we are interested in identifying fake news by testing location consistency -- whether an image is taken at a claimed location. Here, a claimed position could be inferred or obtained from different sources in a social media post, \eg{} associated text description, Global Positioning System (GPS) information in image metadata, scene text in an image like street signs/landmark names, etc. 

A straight-forward solution to this problem is to use the GPS estimation approach, which estimates a query image's GPS according to visually similar images with known GPS locations in a large, geo-tagged reference database and compares the estimated GPS to the claimed one to make the decision. Depending on the used features, one may further classify existing approaches into two families: (1) 2D-only, which uses image features~\cite{lowe2004distinctive,chen2011city,lee2017fast,lefevre2017toward,majdik2013mav,nister2006scalable,arandjelovic2014dislocation} \eg Invariant Feature Transform (SIFT), Speeded-Up Robust Features (SURF) ~\cite{lowe2004distinctive,chen2011city,lee2017fast,lefevre2017toward}, bag-of-words representation~\cite{majdik2013mav}, and vocabulary trees~\cite{nister2006scalable}, to efficiently and effectively retrieve visually similar images first, and estimate the query GPS from nearest neighbors; 2) 2D-to-3D ~\cite{torii2011visual,sattler2012improving,torii201524,sattler2017efficient}, which reconstructs 3D structures of images in a reference database offline, and performs online 2D-to-3D matching for a query. 

Unfortunately, this approach does not fit well in the context of image-to-GPS verification for three reasons. First, the premise of a large enough, up-to-date, offline reference database is difficult to achieve for most users because such a database is too expensive to create or maintain. Second, we only have one query image instead of a collection or a sequence of images, and thus violate the working assumptions of methods like ~\cite{agarwal2015metric,weyand2016planet}. Third, similarity-based retrieval works well for city landmarks, but not for visually similar locations, \eg{} \emph{Starbucks} stores in different places all over the world. 

Alternatively, we approach this problem following the classic image verification paradigm -- given a pair of images, one query and one reference, we use a network to decide whether or not they are from the same location, where the reference image can be retrieved at the claimed GPS location from a third-party GPS-based image database \eg{} \emph{Google Street View}~\cite{anguelov2010google} and \emph{Bing Street Side}. 
Of course, many existing works on image verification, \eg{} face verification~\cite{abdalmageed2016face} and object verification~\cite{koch2015siamese}, can be directly applied to this problem because verification nature does not change, but they are unsuitable since the critical camera information like shooting angle and focal length is unknown and this raises difficulty to retrieve an appropriate reference image to compare against the query. The potential mismatch roots in the fact that a query image is a 2D projection of a 3D scene, while a GPS location is a 2D point. 

In this paper, we propose a novel Bottom-Up Pattern Matching (BUPM) based verification network. It directly compares a query image and a panorama reference image collected from a claimed GPS location, and thus completely get rid of the error-prone reference images caused by unknown shooting angle and focal length and largely simplifies the data preparation. It estimates the potential matched patches in both reference and query in a bottom-up manner and makes the decision upon the number of matched patches in a soft way. All modules in the BUPM network are therefore differentiable and learnable. In this way, the BUPM network can be used not only for verification but also for localization, \ie{} finding the query image in a panorama reference image.

The remainder of this paper is organized as follows: Sec.~\ref{sec.related} briefly reviews recent related works; Sec.~\ref{sec.method} introduces the image-to-GPS verification problem and proposes the BUPM verification network solution; Sec.~\ref{sec.training} discusses the details of training and dataset; Sec.~\ref{sec.expt} compares performances of different verification solutions; and we conclude this paper in Sec.~\ref{sec.conclude}.

\section{Related Works}\label{sec.related}
Our problem is closely related to the works in the following two domains: 1) image verification, which answers whether a query image is the \emph{same} as the reference in some sense of interest, \eg{} scene and landmark, and 2) template matching, which finds a template region in a host image.

Recent advances in location/landmark verification or retrieval, are mainly from deep neural networks~\cite{melekhov2016siamese,weyand2016planet,tolias2016rmac,arandjelovic2016netvlad,zhou2014learning,noh2017large,hoang2017selective,vo2017revisiting,schonberger2016vote}. \cite{melekhov2016siamese} proposed a generic Siamese network for landmark image matching. \cite{weyand2016planet} quantified 16 millions of images with geo-tags into 26K regions bins, and trained a classification network to predict the region bin of an image belonging to. 
\cite{arandjelovic2016netvlad} proposed a generalized VLAD layer to improve classic feature pooling layers for place recognition. \cite{zhou2014learning} introduced a 7 million scene-centric database and the so-called \emph{Places-CNN} for place/scene classification. 
\cite{noh2017large} trains a network but to classify city landmarks, but introduces a new attention layer to provide patches importance and supervision to fuse all patch features into an image feature. \cite{hoang2017selective} showed that extra attention maps, \eg the density map of SIFT keypoints, helped feature aggregation. \texttt{IM2GPS}\cite{vo2017revisiting} indexed six million geo-tagged images and estimated a query image's GPS via nearest neighbor search using DNN features. 

With regards to template matching, \cite{dekel2015best} proposed the \emph{best-buddies} similarity score, \cite{talmi2017template} used the diversity of feature matches as a similarity metric, and \cite{korman2018oatm} introduced an occlusion aware template matching technique. \cite{wu2017deep} worked on a constrained image splicing localization problem, a general template matching problem, and proposed a learnable deep matching layer to perform matching. 

In the context of the image-to-GPS verification problem, a perspective looking query image and a panorama reference image are not directly comparable. A natural solution is to integrate template matching and image verification into a single solution, and this is exactly what the proposed BUPM network is aiming to achieve. 



\section{The BUPM Network for Image-to-GPS Verification}~\label{sec.method} 
\subsection{Problem Description and Method Overview}
The image-to-GPS verification problem can be stated as follows: given a query image $Q$ and a GPS location $(lat.,long.)$, how to verify whether or not $Q$ is taken at the claimed location. As mentioned before, we treat it as an image verification problem -- whether the visual content of query image $Q$ can be seen in a reference image $R$ retrieved at the claimed location $(lat.,long.)$ through a third-party database, \eg{} \emph{Google Street View}. 

Due to the 2D point nature of a GPS location, we have two choices for the reference image: 1) use many reference images retrieved at the claimed GPS location with different heading angles and focal lengths, verify each one against the query, and aggregate all results for a final decision; and 2) use a panorama reference image, which stitches all scenes spanning over 360 degrees at the claimed GPS location, and verify only this single panorama reference image against the query. 

\begin{figure}[!h]
    \centering
    \setlength{\belowcaptionskip}{-10pt}
    \includegraphics[trim=3cm 2.1cm 2cm 2.5cm, clip, width=0.8\linewidth]{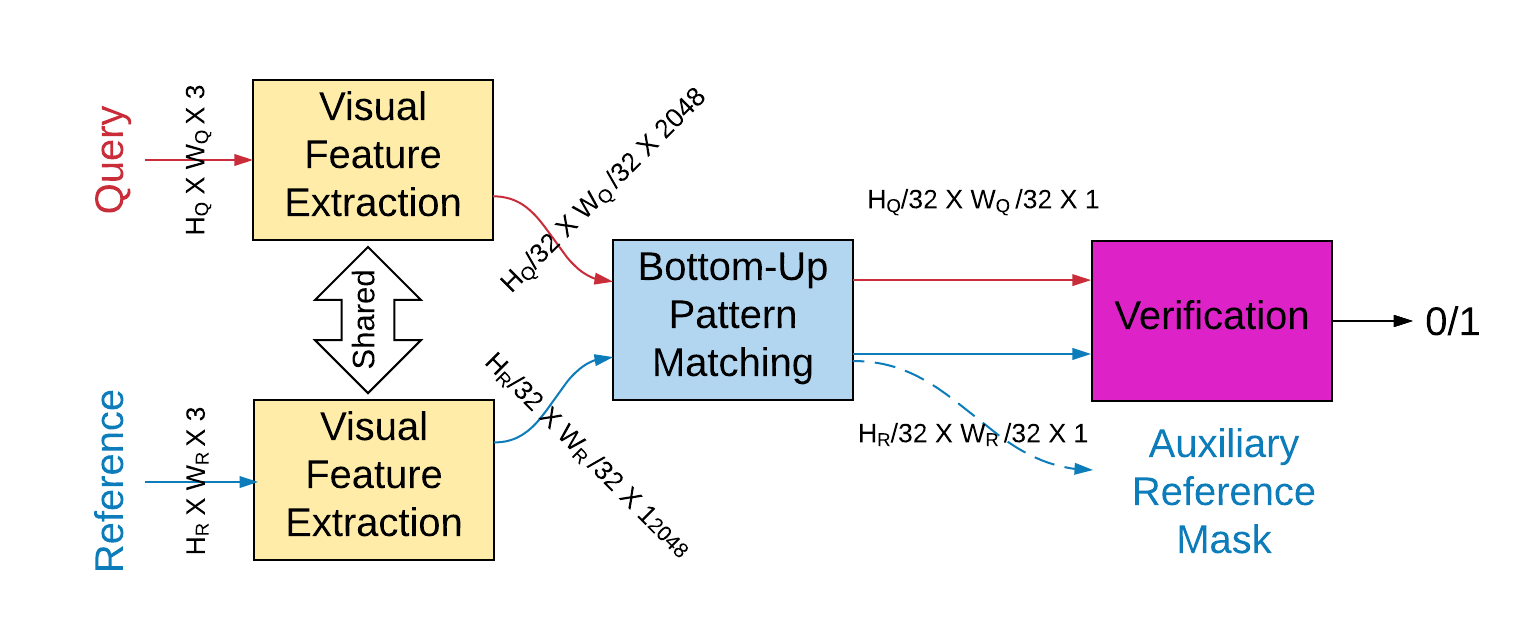}
    \caption{Overview of the proposed bottom-up pattern matching network. Solid arrows indicate the main network (see Sec.~\ref{sec.method}); and dashed arrows indicate the auxiliary network to stabilize the main network (see Sec.~\ref{sec.training}).}
    \label{fig:overview}
\end{figure}

\noindent In this paper, we choose to use the later one to avoid the annoying and error-prone reference image retrieval for the unknown shooting angle and focal length, and the inefficient use of many reference images. However, as a trade-off, the later choice introduces two additional challenges we have to face: 
\begin{itemize}
    \item visual distortions in a panorama image.
    \item extra but irrelevant scene content other than the required.
\end{itemize}

\begin{figure}[!h]
    \centering
    \setlength{\belowcaptionskip}{-25pt}
    \begin{tabular}{ccp{.2cm}|p{.2cm}cc}
        \multicolumn{2}{c}{Sample ($Q$,$R$) Pair A} &&& \multicolumn{2}{c}{Sample ($Q$,$R$) Pair B}\\
         \includegraphics[width=0.11\linewidth]{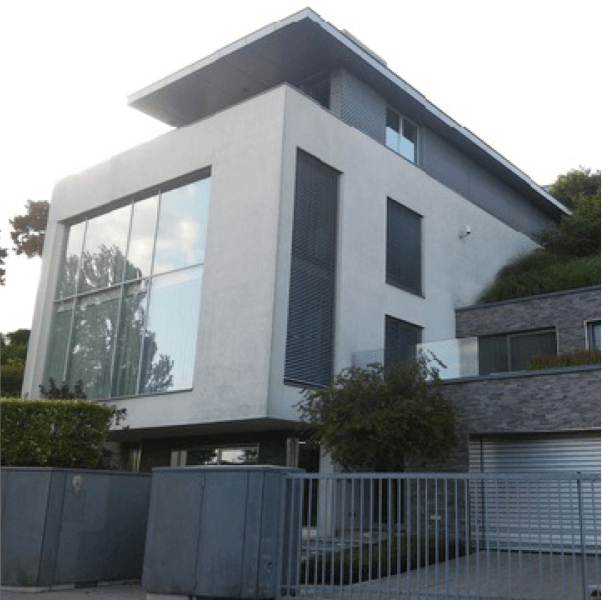}&  
         \includegraphics[width=0.33\linewidth]{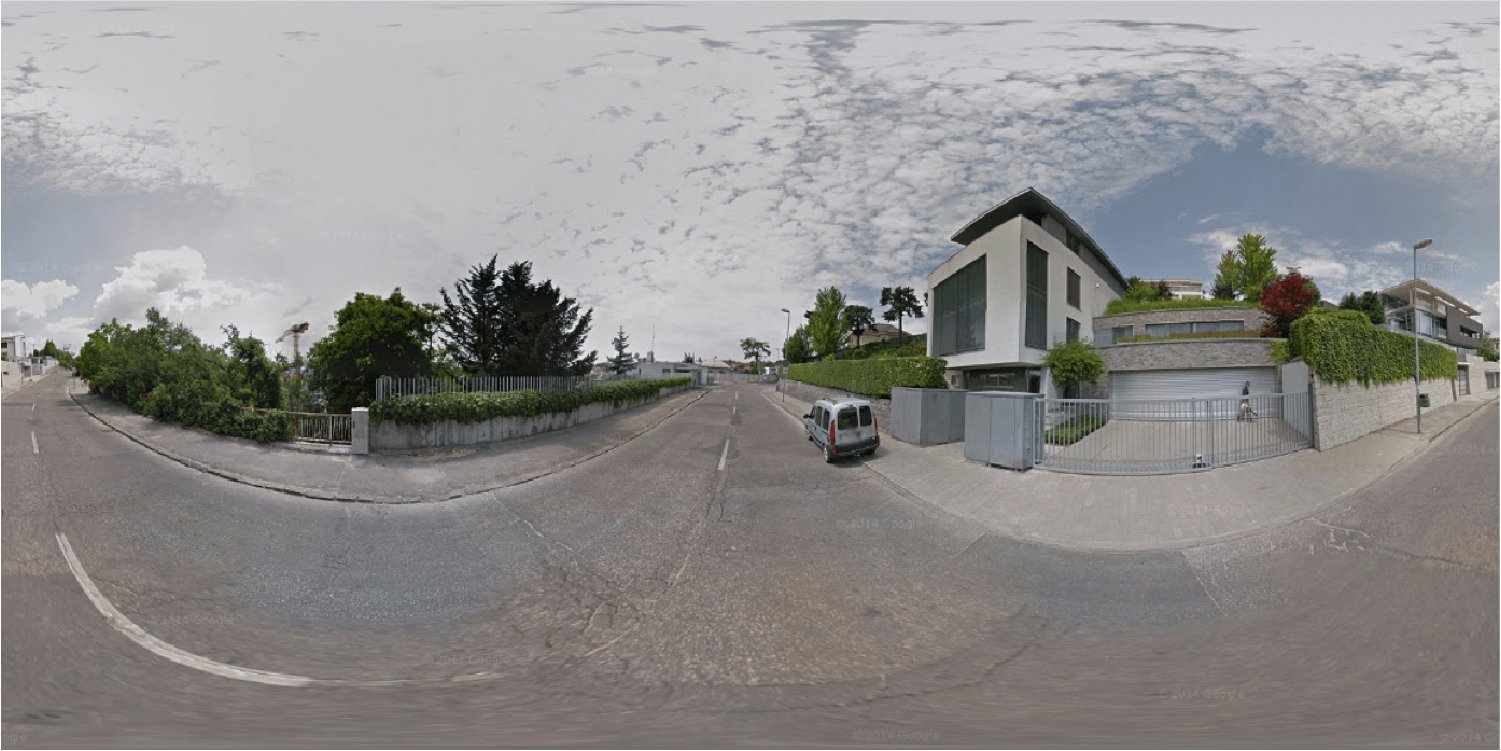}&&&
         \includegraphics[width=0.11\linewidth]{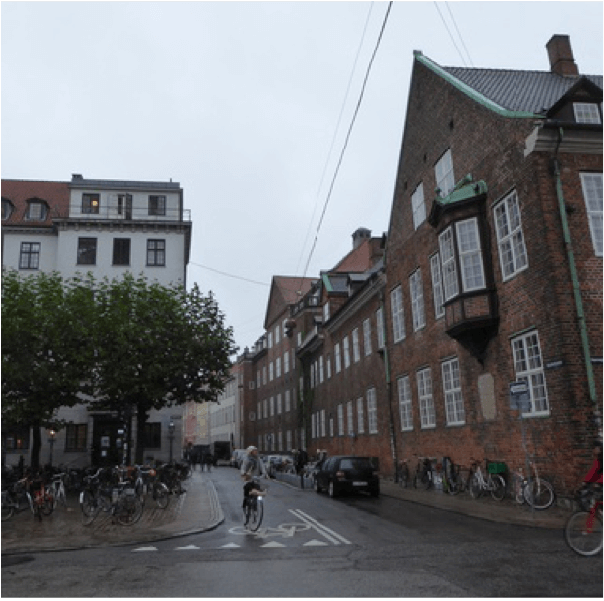}&  
         \includegraphics[width=0.33\linewidth]{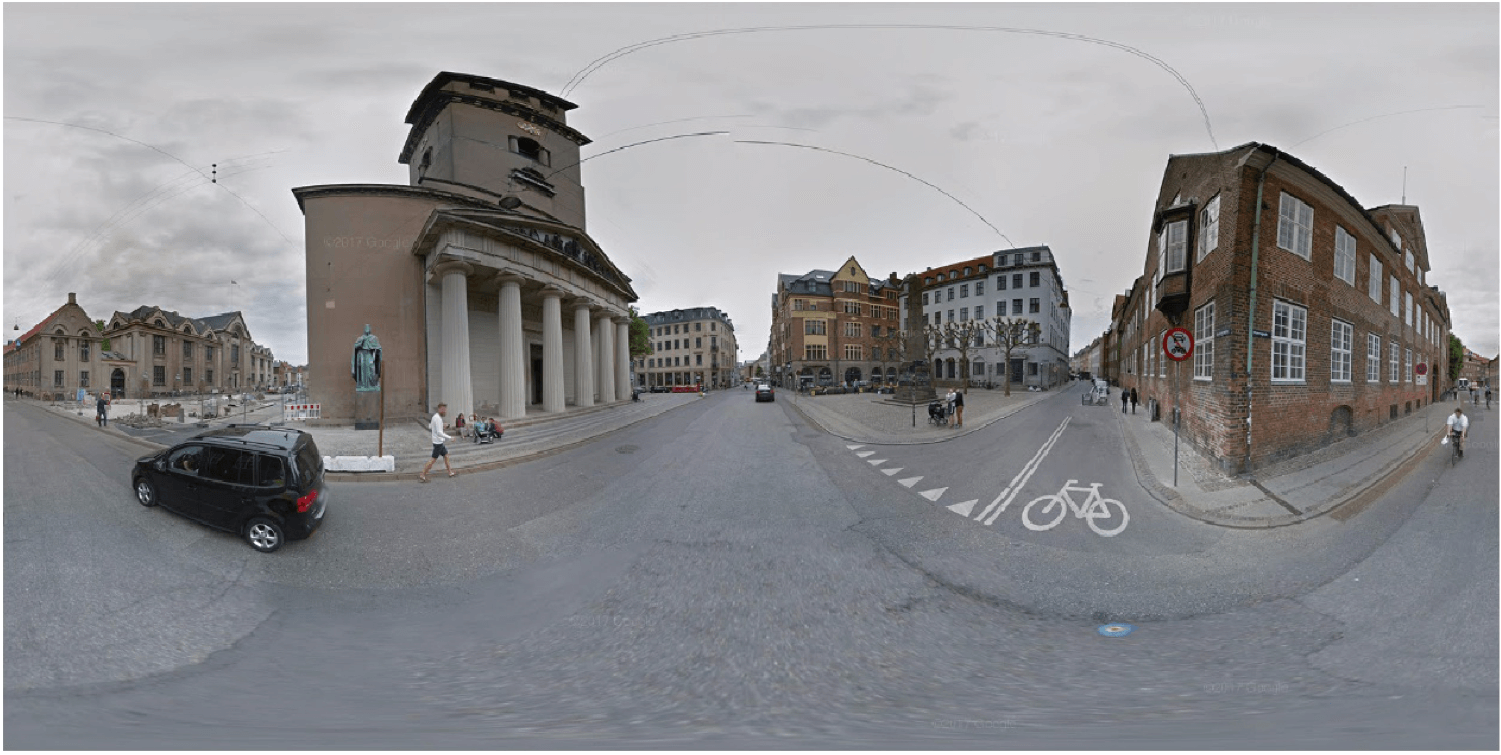}\\
         \includegraphics[width=0.11\linewidth]{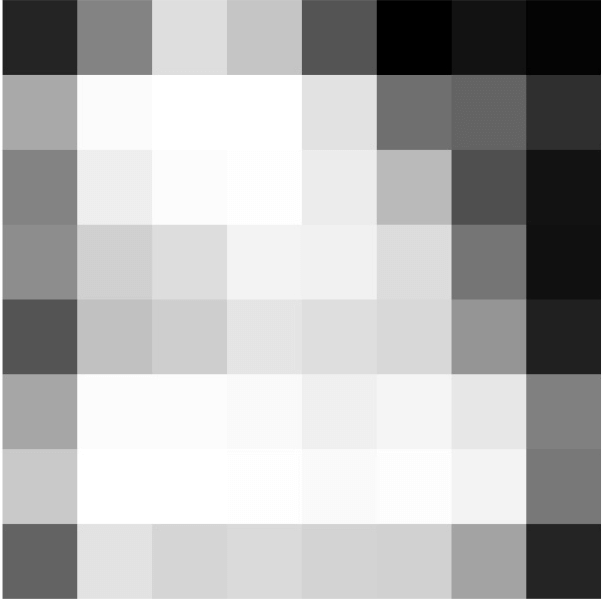}&  
         \includegraphics[width=0.33\linewidth]{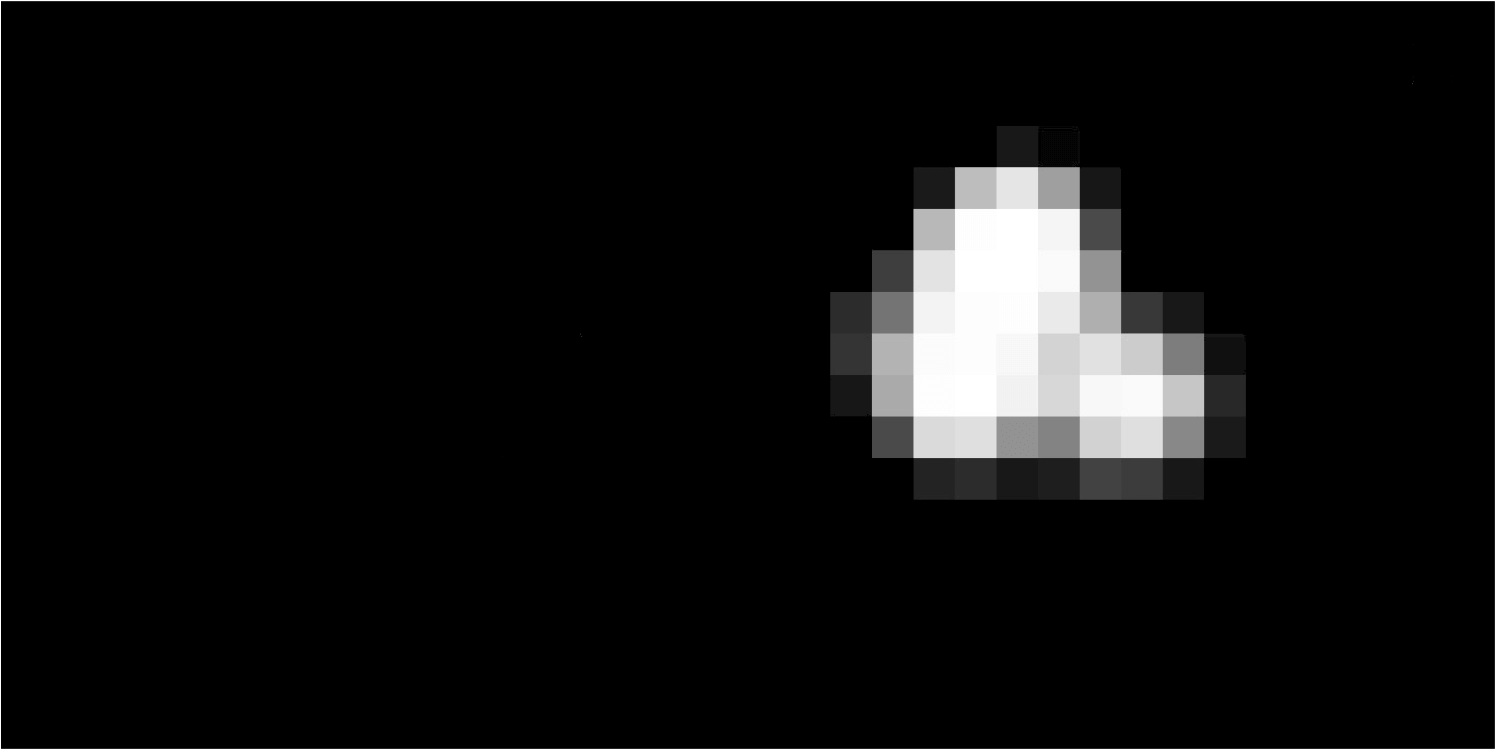}&&&
         \includegraphics[width=0.11\linewidth]{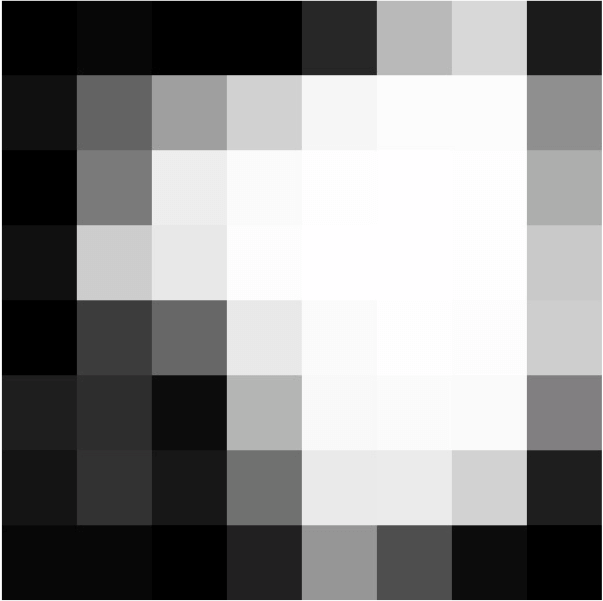}&  
         \includegraphics[width=0.33\linewidth]{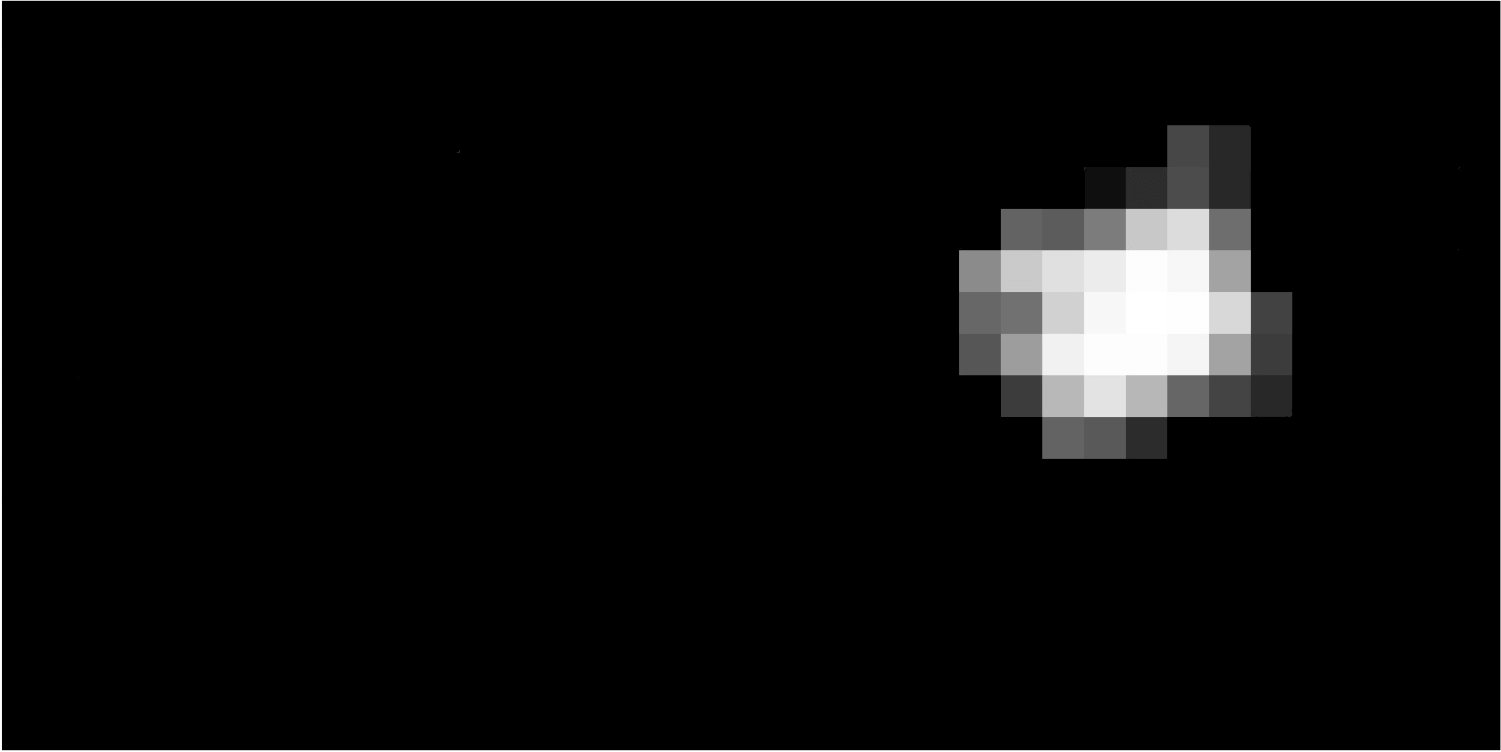}\\    
         \includegraphics[width=0.11\linewidth]{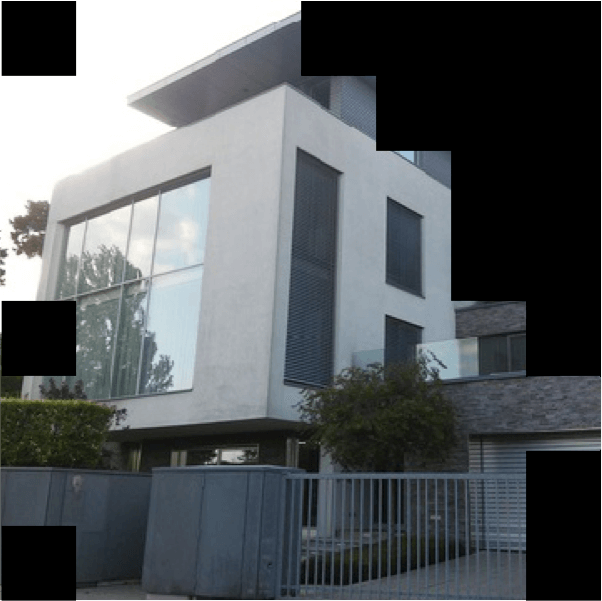}&  
         \includegraphics[width=0.33\linewidth]{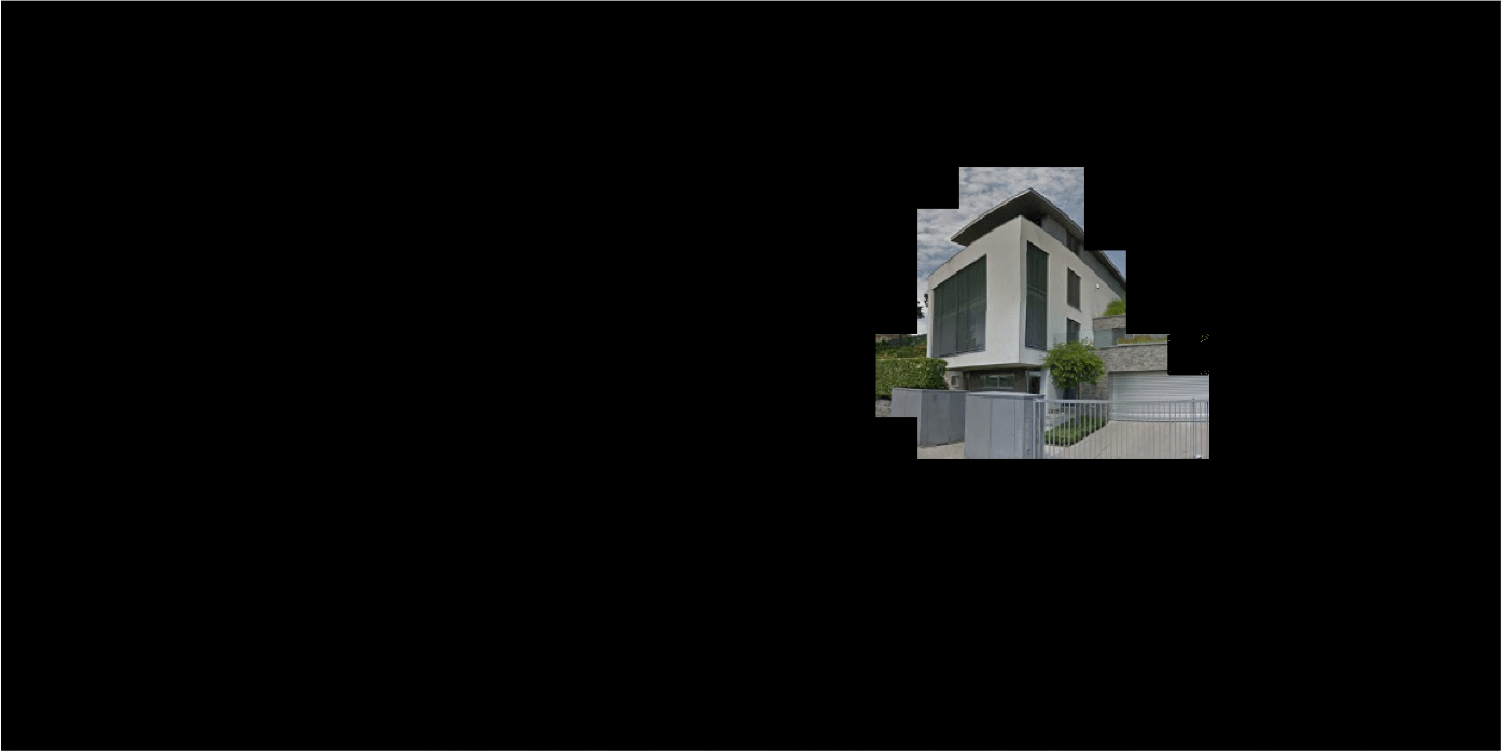}&&&
         \includegraphics[width=0.11\linewidth]{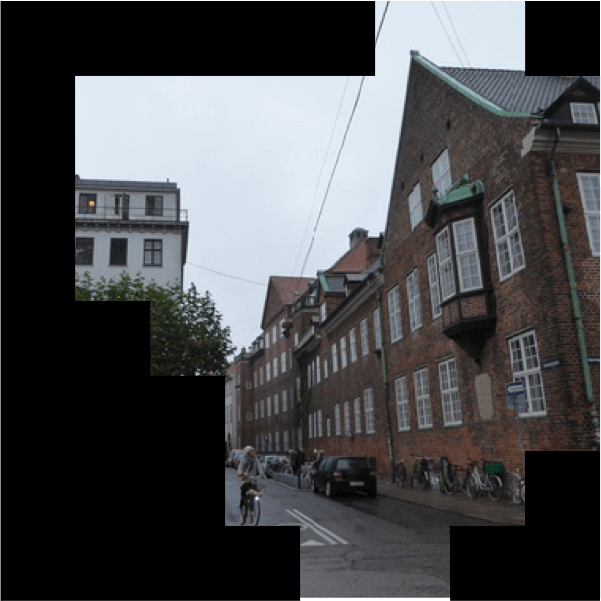}&  
         \includegraphics[width=0.33\linewidth]{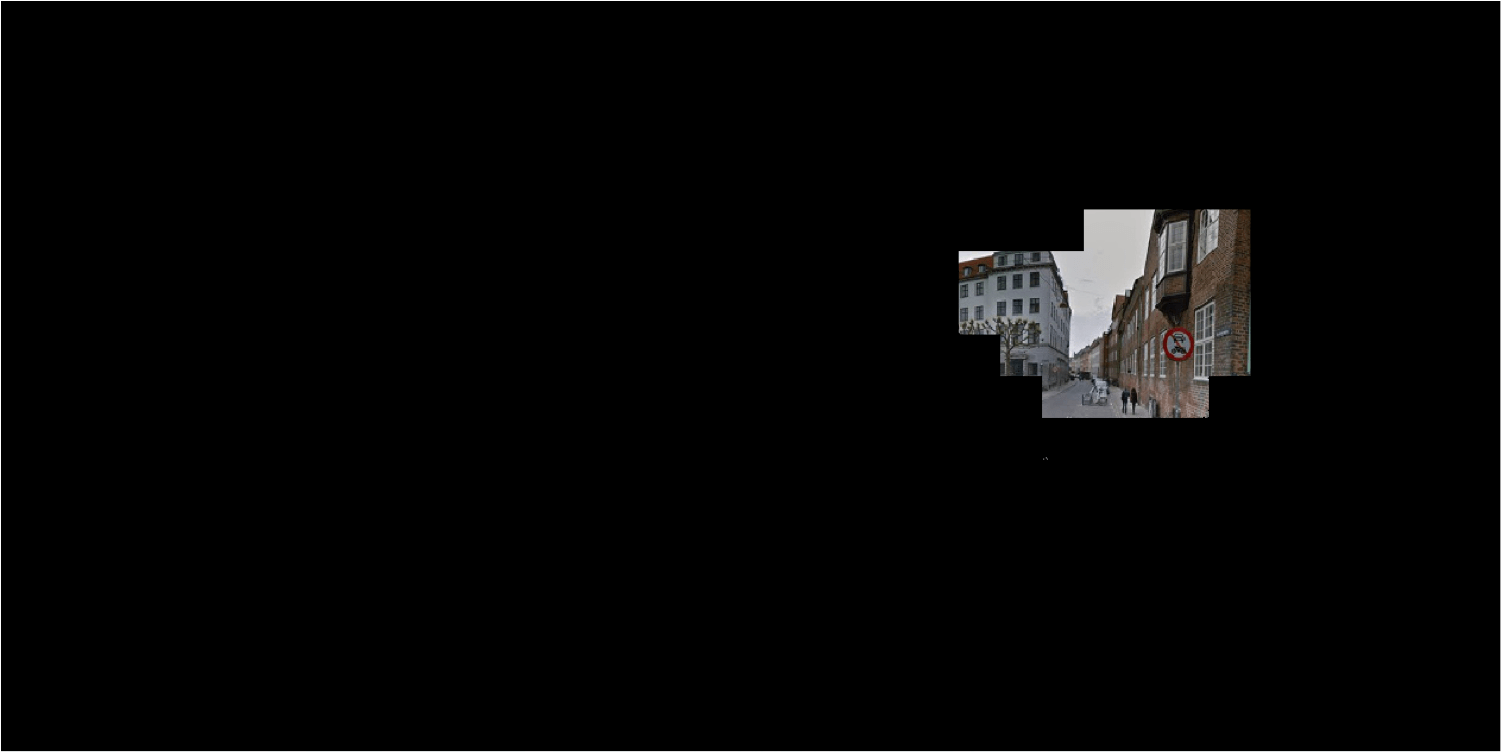}\\ 
    \end{tabular}
    \caption{Sample matching results using the proposed BUPM method. From top to bottom, rows indicate the original input ($Q$, $R$) pairs, the BUPM produced matched region masks, and the overlaid results of the original input and mask (after thresholding).}
    \label{fig:bumpSample}
\end{figure}

Fig.~\ref{fig:overview} shows the overview of our approach. Specifically, the \emph{Visual Feature Extraction} module represents both query and reference images in terms of two feature tensors, the \emph{Bottom-Up Pattern Matching} module finds matched regions between query and reference, and finally the \emph{Verification} module decides whether or not the visual content of $Q$ is seen in the reference $R$ upon the potential matched patches found in query and reference. It is worth noting that the BUPM network works for image pairs of arbitrary sizes. See Fig.~\ref{fig:bumpSample} for sample inputs and BUPM matching results. Details of each module as well as our solutions to the above challenges will be discussed in the following sections.

\subsection{Visual Feature Extraction}\label{sec.pmn}
Although a panorama image contains visual distortions, it is clear that such distortions only exist in fairly large patches. For example, a straight line will be distorted to a curve in a panorama image, but this curve can be still considered to be piece-wise linear when we use small patch sizes. This observation indicates that we don't need to worry too much about distortions in a panorama image if we are only interested in local features.

To extract local visual features, we simply use the pretrained \emph{ResNet50} model~\cite{he2016deep}. More precisely, we crop off all \emph{Dense}(also known as \emph{Fully Connected}) layers and the \emph{Global Pooling} layer in \emph{ResNet50}, while only keeping its \emph{Convolution} layers, because all we need is local visual feature representation while those cropped off layers are for the image classification task. As a result, given a query image of size $H_Q\times W_Q\times3$, and a reference image of size $H_R\times W_R\times3$, the \emph{Visual Feature Extraction} module produces query representation $F_Q$ of size $\frac{H_Q}{32}\times \frac{W_Q}{32}\times2048$, and reference representation $F_R$ of size $\frac{H_R}{32}\times \frac{W_R}{32}\times2048$, where the denominator 32 is caused by the five times of factor 2 downsampling in \emph{ResNet50}, and the depth dimension 2048 is the number of filters used by the last convolutional layer of \emph{ResNet50}. 

\subsection{Bottom-Up Pattern Matching}
 Due to the nature of the panorama image, a reference feature $F_R$ contains much more content than that is required to verify $F_Q$. Therefore, directly comparing $F_R$ and $F_Q$ makes very little sense -- only a small region in $F_R$ is supposed to match with $F_Q$ when $R$ and $Q$ are taken at the same place, while no region should be matched when they are not. Directly comparing these two features means to match signals with the appearance of very heavy noise, and thus its effectiveness and robustness is questionable.
 
\begin{figure}[!h]
    \centering
    \setlength{\belowcaptionskip}{-10pt}
    \includegraphics[trim=6cm 1cm 6cm 1cm, clip,width=.8\linewidth]{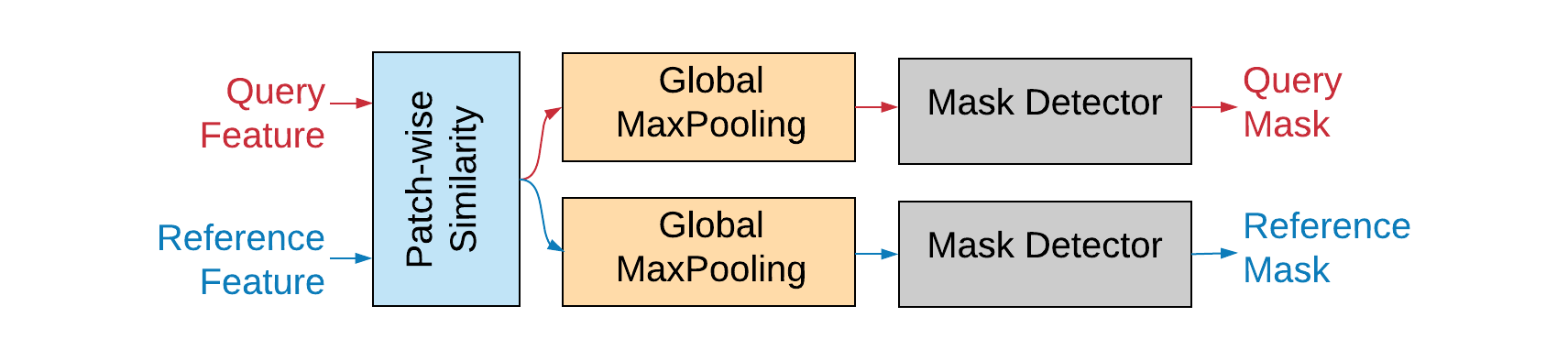}\\{(a) The BUPM module overview}\\
    \includegraphics[width=.9\linewidth]{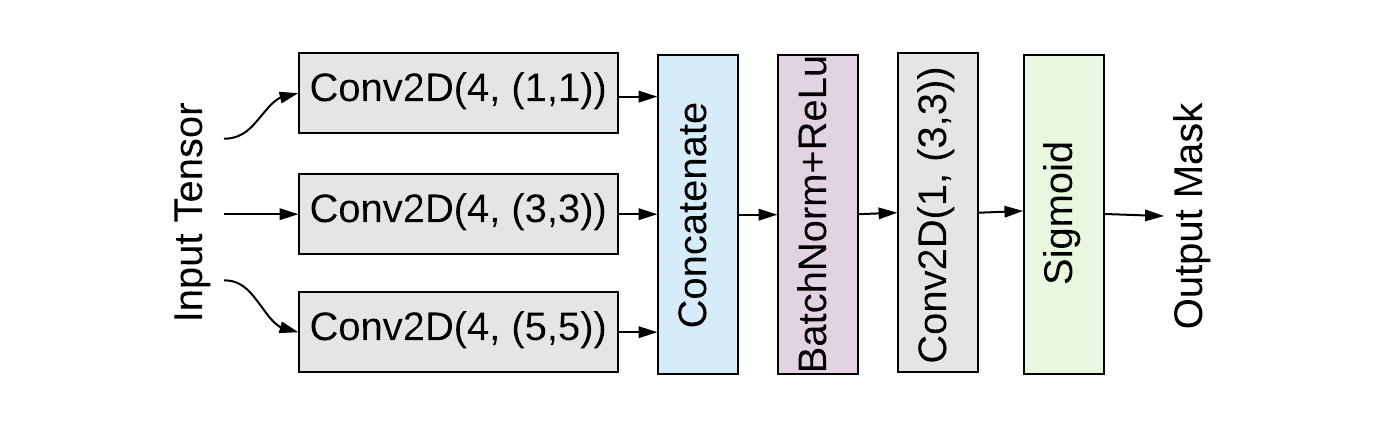}\\(b) The network architecture of the \emph{Mask Detector} module.
    \caption{The proposed bottom-up pattern recognition module.}
    \label{fig:bump}
\end{figure}

Alternatively, we follow the procedure adopted by a human user, who will first look for regions that are similar to a query, and later decide whether the found regions are matched or not. Specifically, the proposed BUPM module consumes query and reference feature tensors and does three things, 
 \begin{enumerate}
     \item computes similarity scores between reference and query from the bottom in the \emph{Patch-wise Similarity} module
     \item find best matched patch pairs through \emph{Global MaxPooling}
     \item holistic matching based mask detection in the \emph{Mask Detector} module.
 \end{enumerate}
This process is illustrated in Fig.~\ref{fig:bump}. 

To match patterns in the bottom-up manner, BUPM views a feature tensor of size $H\times W \times 2048$ as a collection of $H\times W$ patch-wise feature vectors, each of dimension $2048$. Consequently, the smallest unit for analysis is of the patch-level. BUPM then computes a pairwise cosine similarity tensor $S=\textrm{Simi}(F_R, F_Q)$, where each element can be computed as the similarity score between a query patch feature and a reference patch feature as shown in Eq.~\eqref{eq:simi}.
\begin{equation}\label{eq:simi}
    \textrm{Simi}(F_R, F_Q)[x,y,i,j]=\frac{F_R[x,y]\cdot F_Q[i,j]}{\norm{F_R[x,y]} \cdot \norm{F_Q[i,j]} }
\end{equation}
A larger $\textrm{Simi}(F_R, F_Q)[x,y,i,j]$ value therefore indicates that reference patch feature $F_R[x,y]$ and query patch feature $F_Q[i,j]$ are more similar. In addition, $S[x,y,:,:]$ stores similarity scores between reference patch feature $F_R[x,y]$ and all query patch features, and $S[:,:,i,j]$ stores similarity scores between query patch feature $F_Q[i,j]$ and all reference features. 

To see which reference/query patch is matched, we perform the \emph{Global MaxPooling} for both query and reference as shown in Eqs.\eqref{eq:poolRef} and \eqref{eq:poolQry}. As one can see, the resulting $B_R$ is of size $\frac{H_R}{32}\times\frac{W_R}{32}\times 1$, and each of its element $B_R[x,y]$ indicates the best matching score we found between the patch feature $F_R[x,y]$ and any feature in $F_Q$. The resulting $B_Q$ can be interpreted in a similar way. 

\begin{equation}\label{eq:poolRef}
    B_R=\textrm{max}(S, axis=(2,3))
\end{equation}
\begin{equation}\label{eq:poolQry}
    B_Q=\textrm{max}(S, axis=(0,1))
\end{equation}

Once these best matching scores $B_R$ and $B_Q$ are obtained, BUPM applies holistic matching to detect mask $M_R$ and $M_Q$ as shown in Fig.~\ref{fig:bump}-(b). Here, the holistic matching is implemented as an inception~\cite{szegedy2016rethinking} module composed of three \emph{Convolution} layers, all of which have 4 filters but with kernel sizes at 1, 3 and 5, respectively. Since no pooling layer is involved, the output mask is of the same spatial size as input. Sample BUPM inputs and outputs can be seen in Fig.~\ref{fig:bumpSample}.

\subsection{Verification}
Inspired by the simple geometric verification used in the classic SIFT matching~\cite{zhou2017effective}, whose verification decision is made upon the number of matched feature pairs between two images, we propose a simple yet effective approach to verify whether or not the query and reference match. More precisely, we compute the number of matched feature pairs in a soft way in three steps: 
\begin{enumerate}
    \item compute the average of detected masks $M_R$ and $M_Q$, respectively
    \item concatenate them as a two-dimensional feature $V$ as shown in Eq.~\eqref{eq:vector}
    \item learn a multilayer perception (MLP) to make verification decision.
\end{enumerate}
Since the input of the MLP is of dimension two, we simply implement it a shallow DNN composed of three \emph{Dense} layers, which are of 16, 4, and 1 units, and followed by the \emph{Sigmoid} activation. 

\begin{equation}\label{eq:vector}
    V = [\textrm{mean}(M_R), \textrm{mean}(M_Q)]^T
\end{equation}

It is worthy noting that computing the mean of $M_R$ and $M_Q$ is equivalent to computing the summation of $M_R$ and $M_Q$, \ie{} counting the number of matched patched but in a soft and differentiable way.

\section{Training The BUPM Network}~\label{sec.training}
\subsection{Real Training Dataset}
To train the BUPM network, we need both positive and negative pairs of perspective-looking query and panorama reference images. Unfortunately, no public available dataset could provide a sufficient number of positive pairs. We therefore collect data for training. 

It is noticeable that positive pairs are more important, because negative pairs can be easily \emph{synthesized} by disordering the matched pairs. We start our data collection with downloading quality query images with GPS coordinates. Due to the sensitivity of different GPS sensors/receivers/algorithms, not all query images with GPS locations are considered. Here, we only consider those query images taken by the recent smart phones, \eg{} Apple iphone 6, which are 
\emph{typically accurate to within a 4.9m (16ft) radius under open sky}\footnote{\url{https://www.gps.gov/systems/gps/performance/accuracy/}}. As a result, \emph{Mapillary}\footnote{\url{www.mapillary.com}}, a photo sharing website is used as the source of query images, most of whose images are taken by GPS-enabled smart phones and uploaded by users all over the world. Once a query image is obtained, we then download its corresponding panorama image through the \emph{Google Street View} API\footnote{\url{https://github.com/Jam3/extract-streetview}}. 

\begin{figure}[!h]
    \centering
    \setlength{\belowcaptionskip}{-10pt}
    \includegraphics[trim=0cm 2cm 0cm .5cm, clip, width=.7\linewidth]{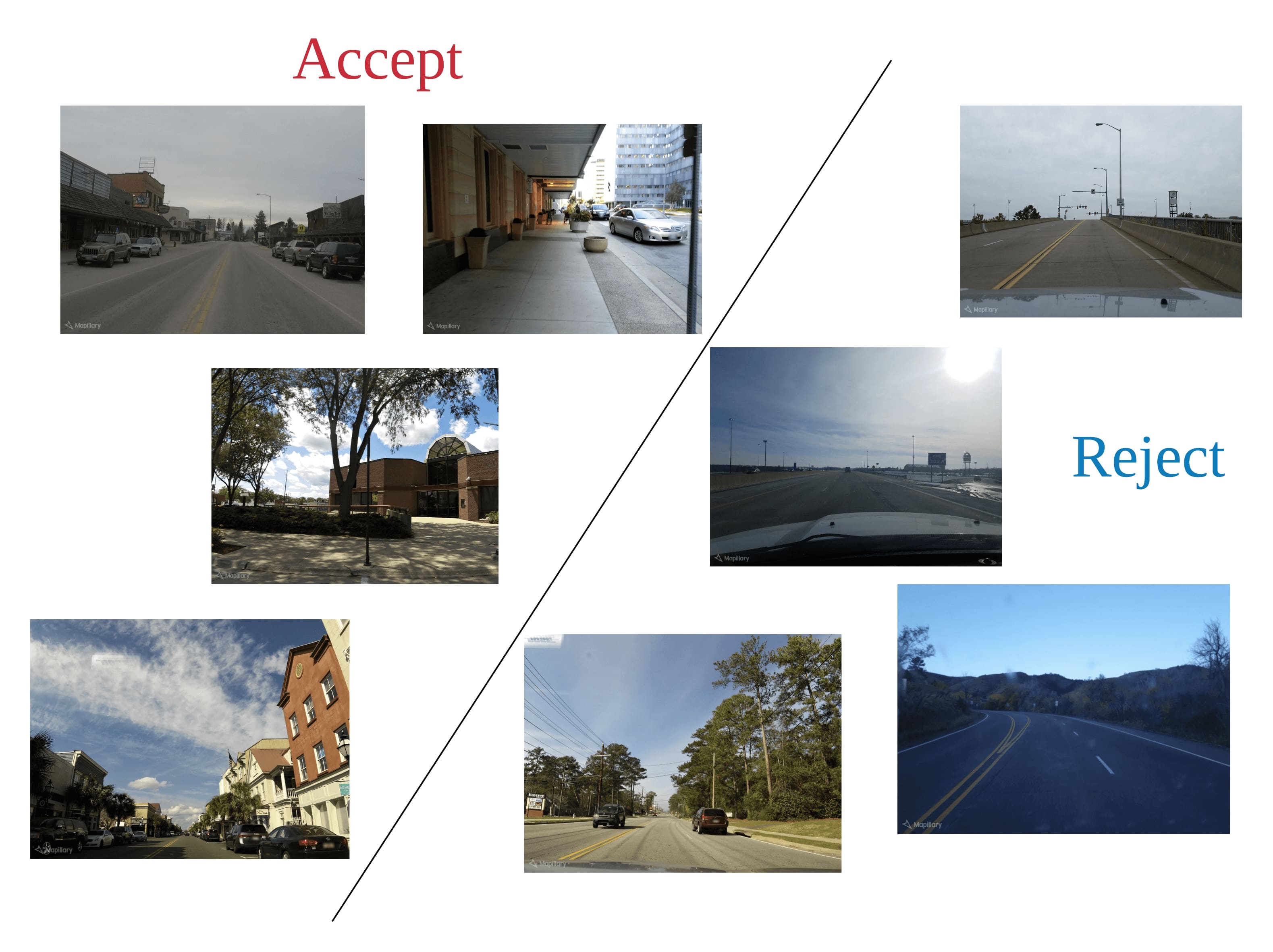}\\
    \caption{Rejected and accepted query instances after data filtering.}
    \label{fig:sampleData}
\end{figure}

In total, we collected $~$120K raw query images and reference panorama images. The raw data are filtered if violating any of the following condition: 
\begin{enumerate}
    \item no panorama image can be retrieved at a GPS (\eg{} indoor images)
    \item no immovable object like buildings in query (\eg{} sky images)
\end{enumerate}

The first violation can be easily identified by checking retrieval return values. The second violation can be detected by using any pretrained semantic segmentation model have building classes, \eg{} the UperNet \footnote{\url{https://github.com/CSAILVision/semantic-segmentation-pytorch}}.  Eventually, we successfully harvested 30K positive samples. Rejected and accepted instances can be seen in Fig.~\ref{fig:sampleData}. We randomly split these 30$K$ of paired query and reference positive samples into training and validation dataset of sizes 25$K$:5$K$. We further disorder paired query and reference samples to obtain 5$K$ negative samples for validation, respectively. For training, we did not use a fixed negative dataset, but randomly generate negative samples for each batch. 

\subsection{Synthetic Training Dataset}

Directly training the proposed BUPM network with the real dataset may fail to converge with a great chance. This is because the network may not optimize towards to the desired directions, \ie{} $M_R$ and $M_Q$ in the proposed BUPM network may not represent the matched regions between query and reference but something different. To enforce this designed feature and stabilize the BUPM training, we prepare a synthetic training dataset with additional $M_R$ targets. 

\begin{figure}[!h]
    \centering
    \setlength{\belowcaptionskip}{-10pt}
    \includegraphics[trim=0cm 0cm 0cm .5cm, clip=True, width=1\linewidth]{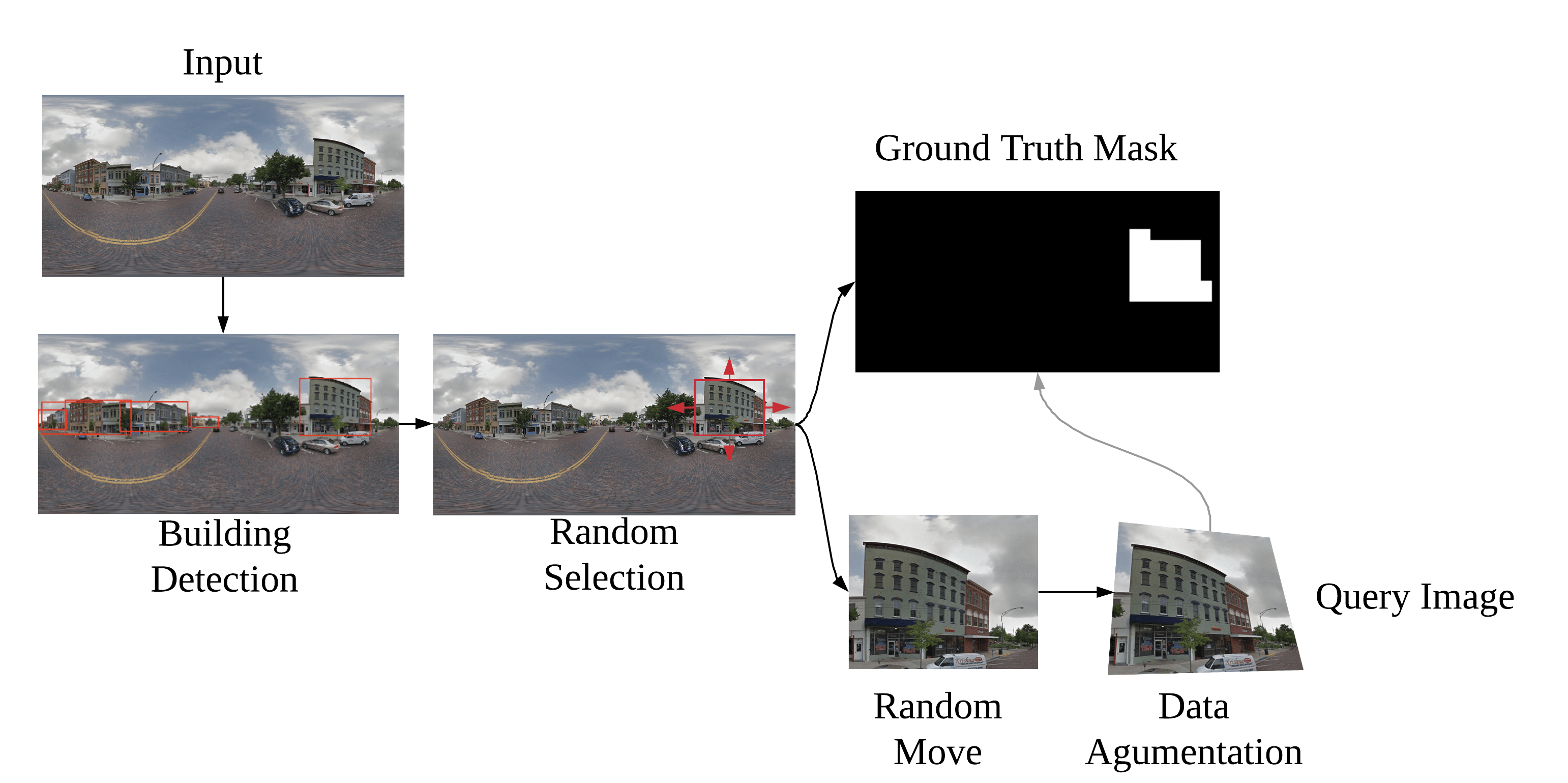}
    \caption{Synthesizing auxiliary \emph{reference masks} and queries for training.}
    \label{fig:synData}
\end{figure}

Specifically, for a given panorama image $R$, we first apply a pretrained building detector to extract a number of region candidates (containing buildings), randomly select one of them, and apply data augmentation to this region to synthesize a query image $Q$. Because we know where $Q$ is taken in $R$, we have $M_R$ targets. This process is described in Fig.~\ref{fig:synData}. The used data augmentation includes scaling (range in $[0.5,2]$), spatial shift (range in (-20\%,20\%)), color adjustment using gamma correction (range in [0.5, 1.5]), and random perspective transform.

\subsection{Training Details}
We implement the BUPM network using the deep learning library \emph{Keras} with the \emph{TensorFlow} backend and 4 \emph{Nvidia Titan-X} GPUs. All weights are randomly initialized except for the \emph{Visual Feature Extraction}, which takes the pretrained \emph{ImageNet} weights from \footnote{\url{https://keras.io/applications/\#resnet50}}. Our pretrained models and dataset can be found in {\color{blue}\url{https://gitlab.vista.isi.edu/chengjia/image-GPS}}.

The training process is composed of two phases: 1) training with the synthetic dataset, and 2) training with the real dataset. In the first phase, we use synthetic query $Q$ and real reference $R$ images to predict $M_R$, \ie{} train the network until the \emph{BUPM} module (see Fig.~\ref{fig:overview}). We use the \emph{logloss} for the $M_R$ target, optimize the network using the \emph{SGD} optimizer with the initial learning rate \emph{1e-2} without any decay, and use a batch size of 16. Note, the $M_Q$ target is omitted, because its ground truth masks in the context of synthetic data are always 1s, and a constant target somewhat hampers the learning. In the second stage, we use the \emph{binary cross-entropy} for the verification target, optimize the network the \emph{Adam} optimizer with the initial learning rate \emph{1e-3}, and set the batch size to 64 with balanced positive and negative samples. Once the MLP in the BUPM network converges, we unfreeze all weights and optimize the BUPM network end-to-end with the \emph{Adam} optimizer with the initial learning rate \emph{1e-5} until convergence.

It is worthy to mention that we resize query images in a batch to the same square size to speed up training. Depending on data augmentation, the query image size can be one of the sizes in 192, 224, and 256.

\section{Experimental Results}~\label{sec.expt}
\subsection{Dataset}
We use two dataset in experiment, namely the \emph{Shibuya} dataset and the \emph{Wikimedia Common} dataset. 
The \emph{Shibuya} dataset~\cite{torii2011visual} is one of very few public dataset with perspective looking query images and panorama reference images. This dataset is originally designed for the location retrieval task, but we reuse all of its 942 query images for our image-to-GPS verification task. It is worthy noting that this is a very challenging dataset in the sense that all reference and query images are densely located in a small geo-region less than 6 square miles but with similar architectures and styles. We pair these 942 query images with 942 panorama images taken at the ground-truth GPS locations to form positive samples, and pair them with additional 942 panorama images that are 1+ miles away from the ground-truth GPS locations to form negative samples. 

\begin{figure}[!h]
    \centering
    \setlength{\belowcaptionskip}{-20pt}
    \includegraphics[width=.6\linewidth]{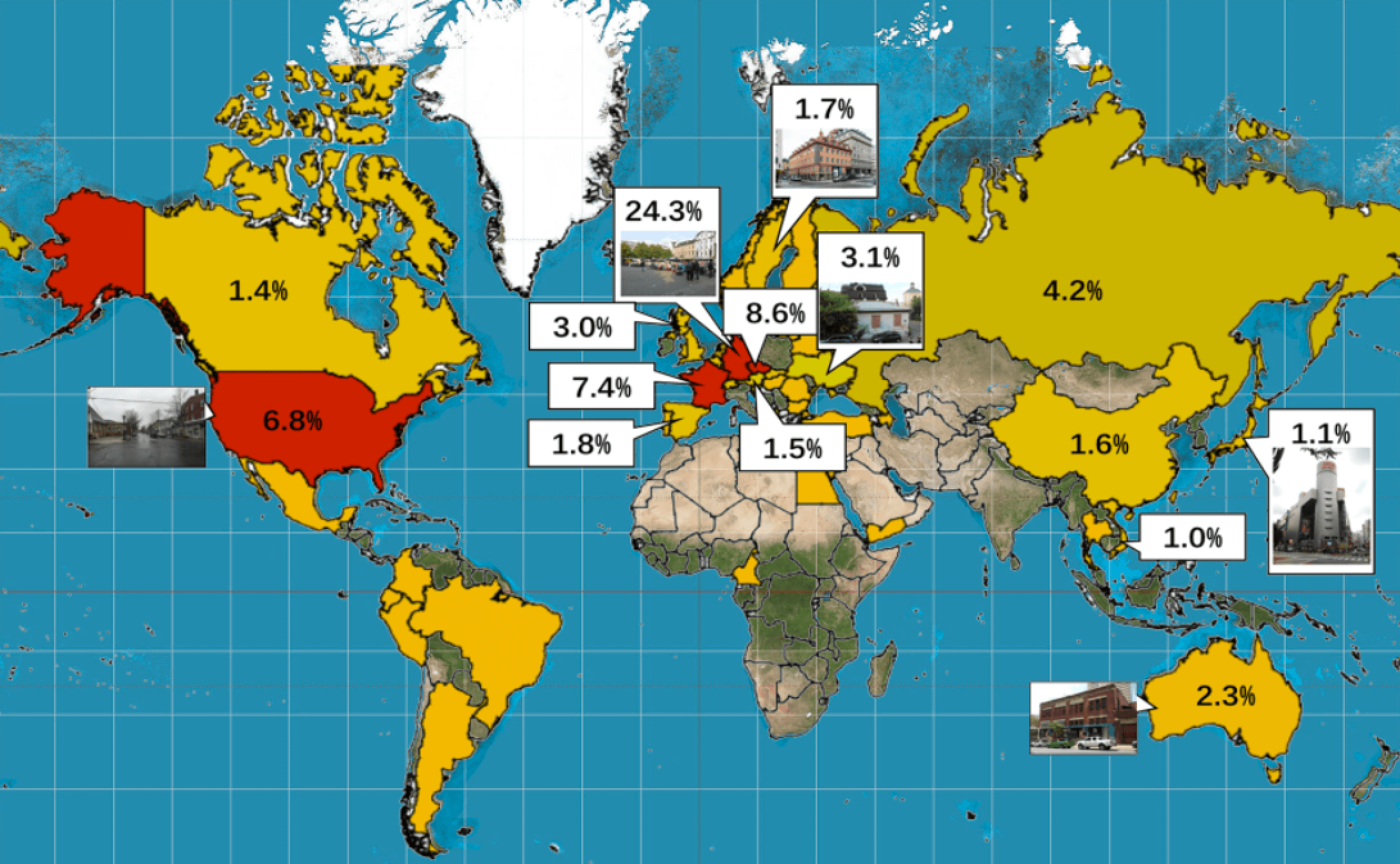}\\
    \caption{The country distribution of the Wikimedia Common dataset.}
    \label{fig:dist}
\end{figure}

The \emph{Wikimedia Common} dataset\footnote{\url{https://commons.wikimedia.org/wiki/Category:Images}} is collected by ourselves for evaluating the BUPM performance on different places. In total, we collected 500 positive samples all over the world, where query images are manually verified to be 1) street-view, 2) taken in recent two years, and 3) visually verified in corresponding panorama reference images. The country distribution of these samples can be seen in Fig.~\ref{fig:dist}. Similar to what we did for the \emph{Shibuya} dataset, we also pair the query images and panorama images to form 500 positive and 500 negative samples. 

\subsection{Baselines}
To understand the performance of the proposed BUPM network, we compare its overall performance with the state-of-the-art location-based image retrieval solutions, and its localization performance with the state-of-the-art template matching solutions. For verification baselines, we use
\begin{itemize}
    \item \emph{NetVLAD}~\cite{arandjelovic2016netvlad} for place recognition.
    \item \emph{Places-CNN}~\cite{zhou2014learning} for place-based scene recognition.
    \item \emph{DELF}~\cite{noh2017large} for location based image retrieval.
\end{itemize}
For template matching baselines, we use
\begin{itemize}
    \item Best-buddies similarity (\emph{BBS})~\cite{dekel2015best}
    \item Deformable diversity similarity (\emph{DDIS})~\cite{talmi2017template}
\end{itemize} 

All pretrained models/methods are directly taken from their repositories or provided by authors. To make these baselines compatible with the proposed image-to-GPS task, we simply treat them as \emph{feature extractor}s, \ie{} cropping-off the last classification layer of the network if necessary, and use the method defined metric to compute feature-wise similarity score between features of a pair of query and reference images. 

\subsection{Metrics and Performance}
To fairly compare method performance while avoiding additional post-processing, we use 1) \emph{Area Under the Curve} (AUC), and 2) \emph{Precision-Recall} plot. Both are operated on various threshold settings. 

\begin{figure}[tb]
    \centering
    \setlength{\belowcaptionskip}{-10pt}
    \scriptsize
    \def\fs{.4}
    \begin{tabular}{@{}c@{}c@{}}
         \includegraphics[ width=\fs\linewidth]{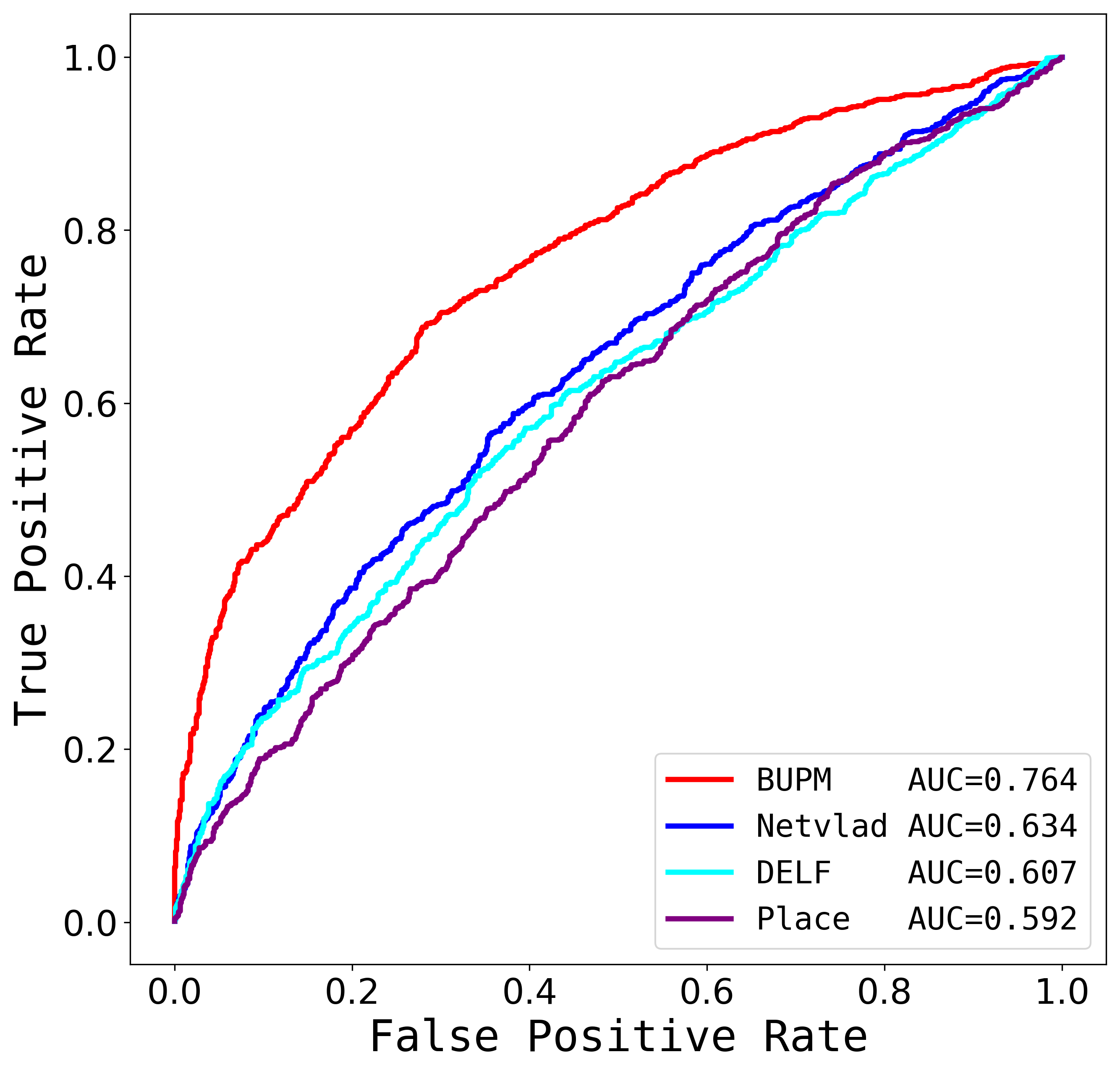}& \includegraphics[width=\fs\linewidth]{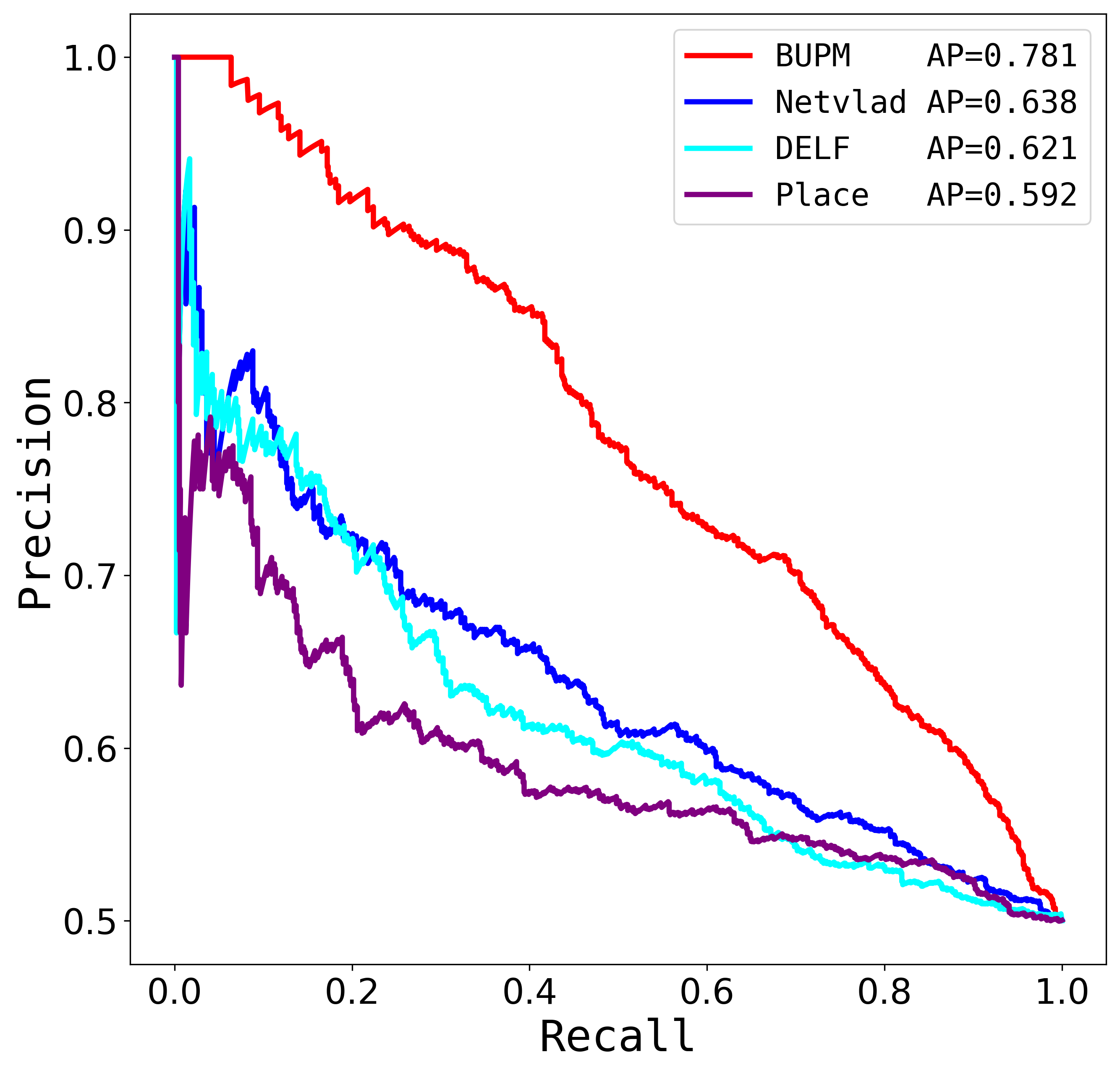} \\
         \multicolumn{2}{c}{(a) Performance on the \emph{Shibuya} dataset} \\
         \includegraphics[width=\fs\linewidth]{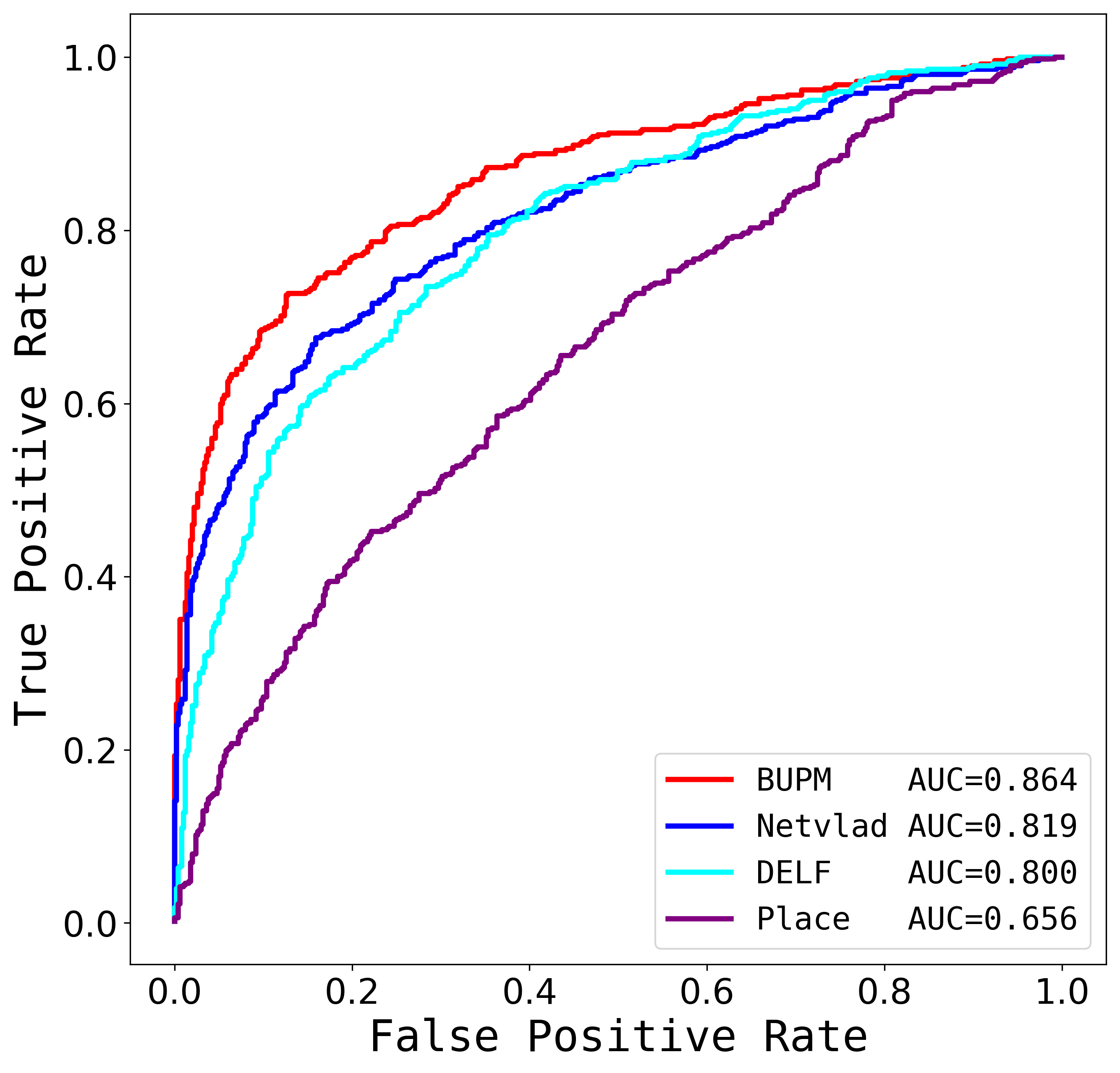}& \includegraphics[width=\fs\linewidth]{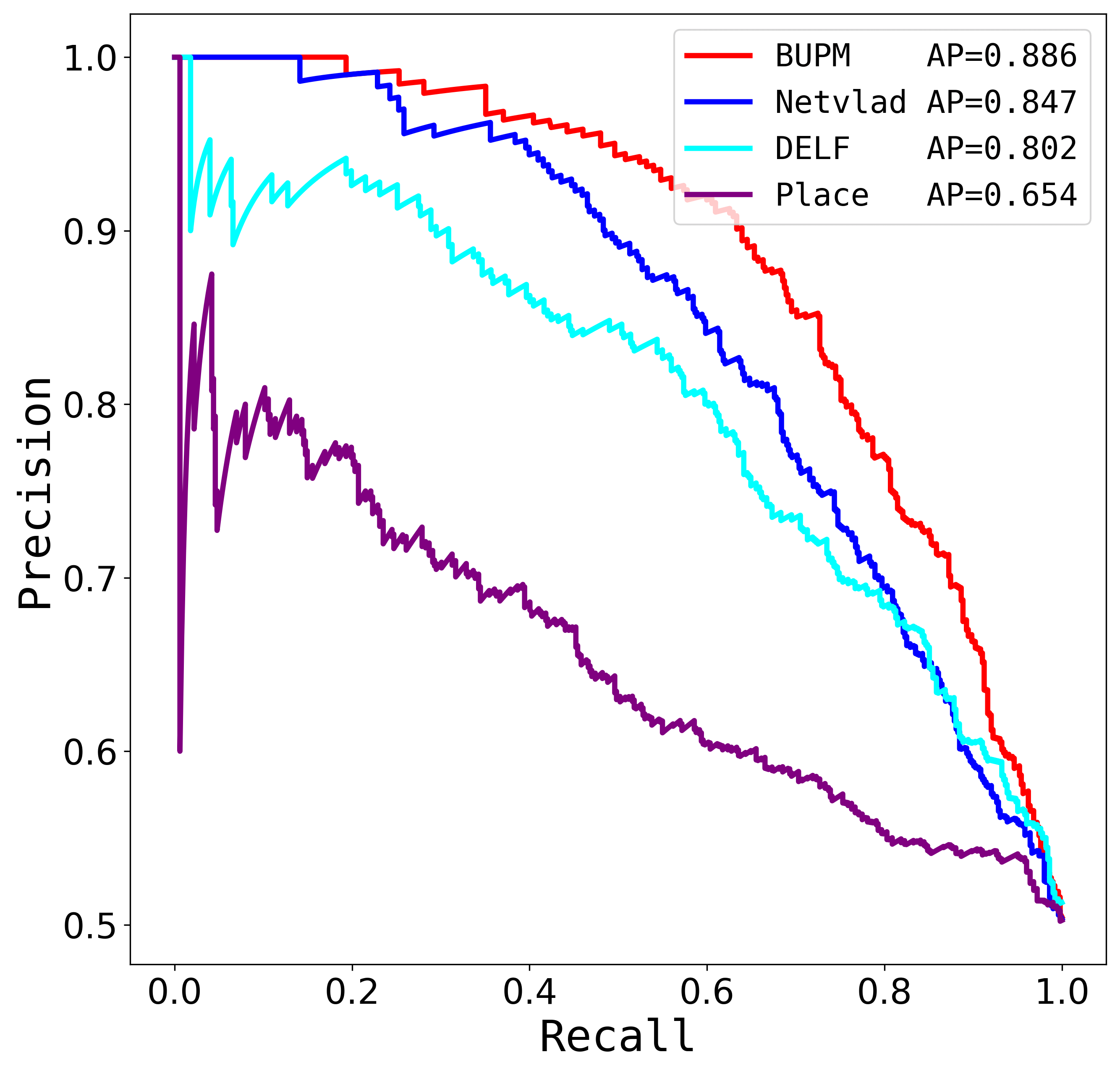} \\
         \multicolumn{2}{c}{(b) Performance on the \emph{Wikimedia Common} dataset} \\
    \end{tabular}
    \caption{Method performance comparisons using the AUC (left) and precision-recall (right) curves. }
    \label{fig:wiki}
\end{figure}

Fig.~\ref{fig:wiki} show the AUC and precision-recall curves of all baselines and the proposed BUPM method for the \emph{Shibuya} and \emph{Wikimedia Common} dataset, respectively. It is clear that the proposed BUPM solution outperforms the state-of-the art solutions by a large margin, leading the second best approach by 13\% (AUC score) and 14\% (average precision score) on the \emph{Shibuya} dataset, and by 4\% (AUC score) and 4\% (average precision score) on the \emph{Wikimedia Common} dataset. The superiority of BUPM is not surprising, because a panorama image contains more contents than required while only the proposed BUPM solution could actively \emph{ignore} these contents.

\begin{figure}[p]
    \centering
    \def\fs{0.95}
    \def\mysquare{$\blacksquare$}
    \scriptsize
    \begin{tabular}{cp{.8cm}cp{.9cm}cp{1.2cm}cp{1.2cm}c}
         \textbf{Query}&& \textbf{Reference} &&\textbf{BUPM} && \textbf{BBS} && \textbf{DDIS}  \\
    \end{tabular}
    \includegraphics[trim=.5cm 5cm .5cm 5cm, clip, width=\fs\linewidth]{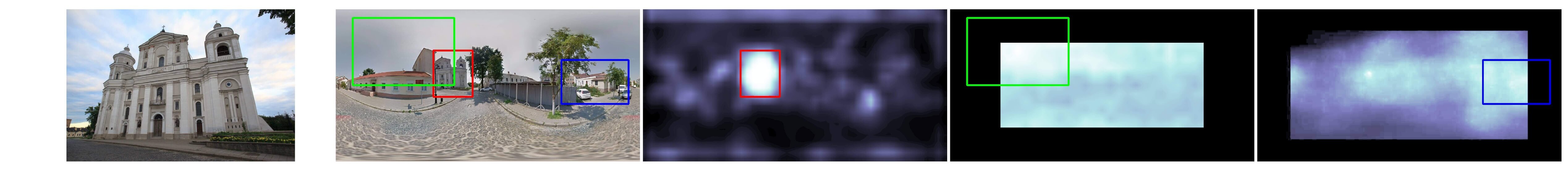}{(a)}\\
    \includegraphics[trim=.5cm 5cm .5cm 5cm, clip, width=\fs\linewidth]{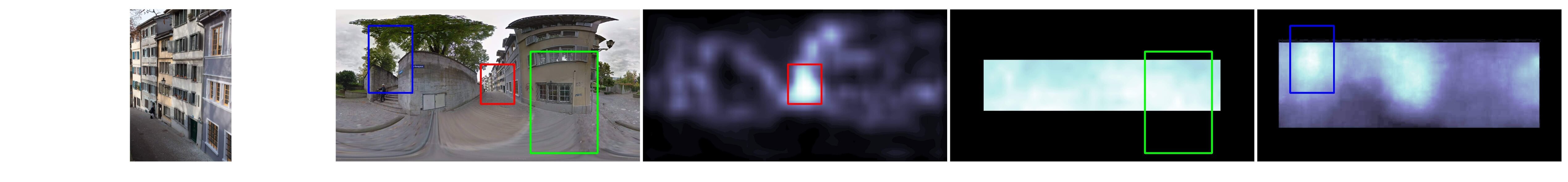}{(b)}\\
    \includegraphics[trim=.5cm 5cm .5cm 5cm, clip, width=\fs\linewidth]{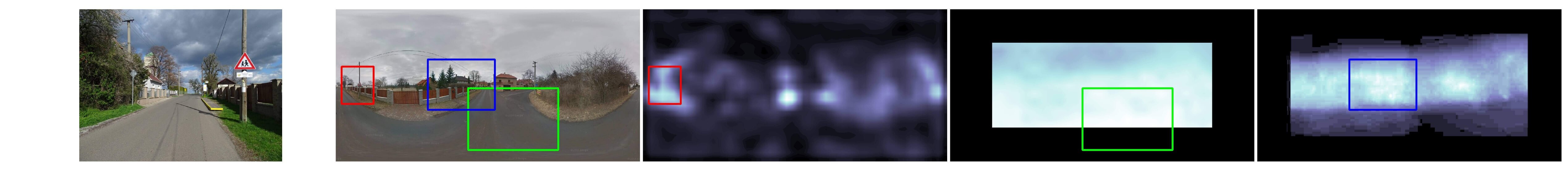}{(c)}\\
    \includegraphics[trim=.5cm 5cm .5cm 5cm, clip, width=\fs\linewidth]{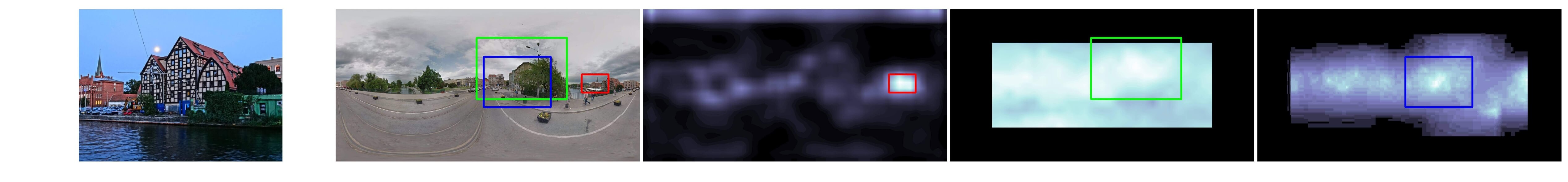}{(d)}\\
    \includegraphics[trim=.5cm 5cm .5cm 5cm, clip, width=\fs\linewidth]{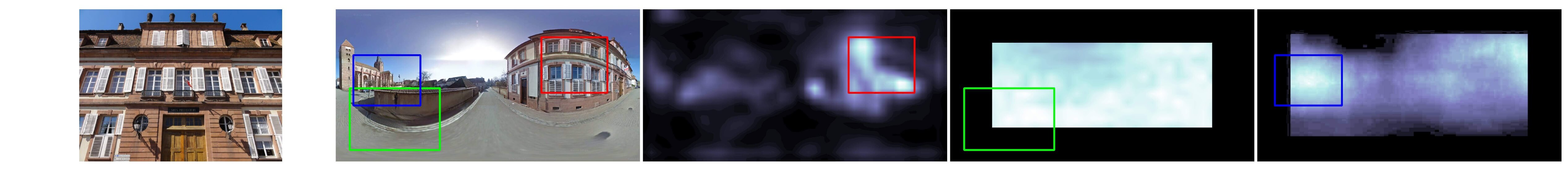}{(e)}\\
    \includegraphics[trim=.5cm 5cm .5cm 5cm, clip, width=\fs\linewidth]{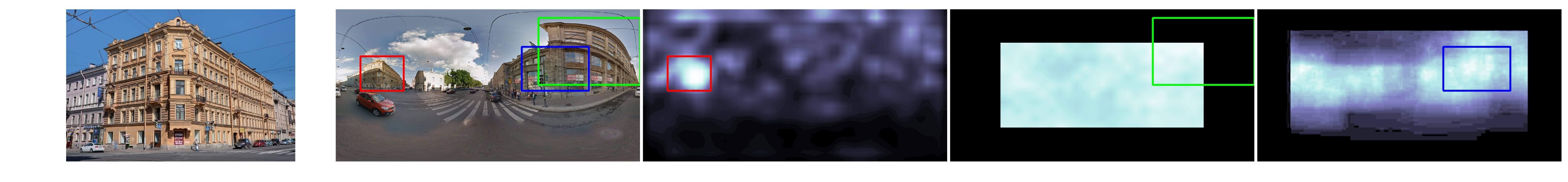}{(f)}\\
    \includegraphics[trim=.5cm 5cm .5cm 5cm, clip, width=\fs\linewidth]{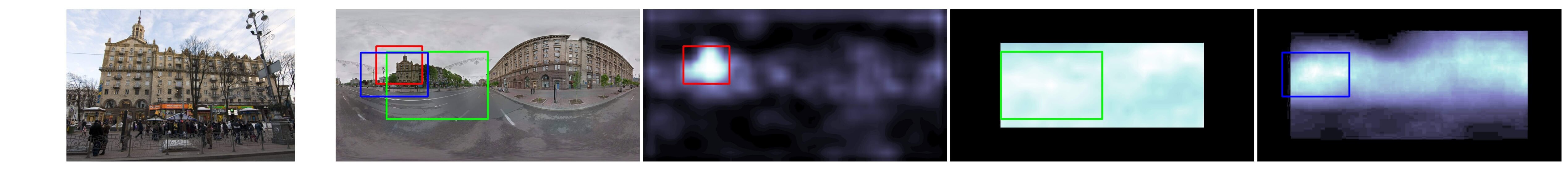}{(g)}\\
    \includegraphics[trim=.5cm 5cm .5cm 5cm, clip, width=\fs\linewidth]{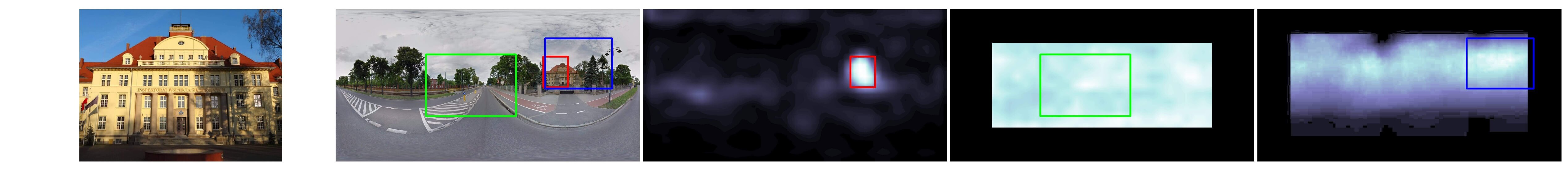}{(h)}\\
    \includegraphics[trim=.5cm 5cm .5cm 5cm, clip, width=\fs\linewidth]{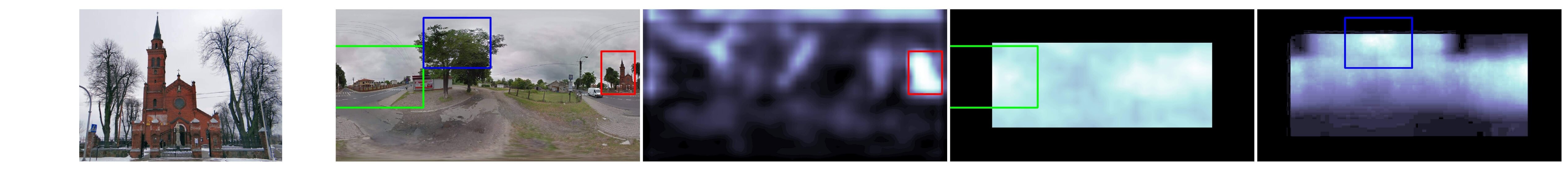}{(i)}\\
    \includegraphics[trim=.5cm 5cm .5cm 5cm, clip, width=\fs\linewidth]{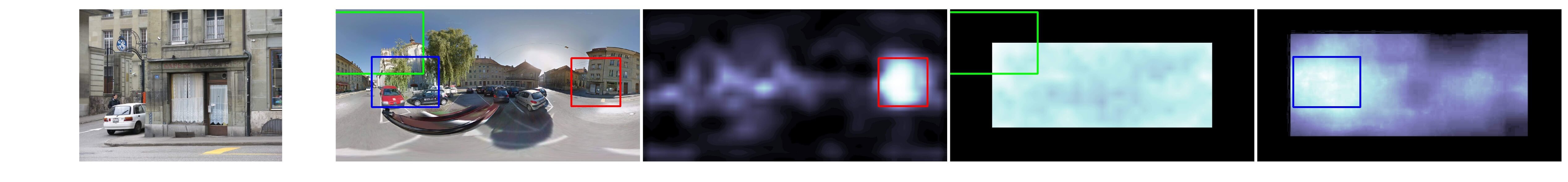}{(j)}\\
    \includegraphics[trim=.5cm 5cm .5cm 5cm, clip, width=\fs\linewidth]{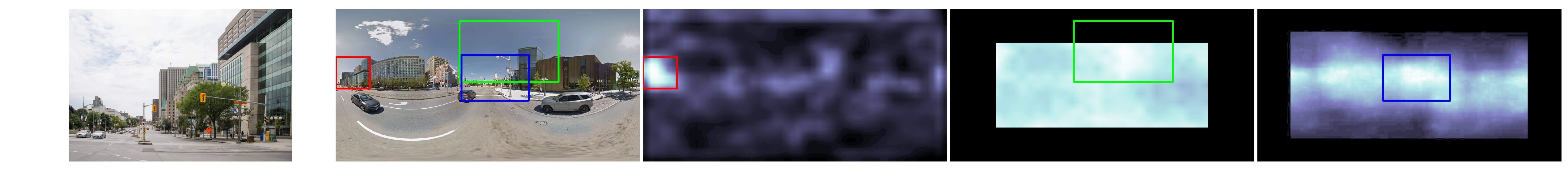}{(k)}\\
    \includegraphics[trim=.5cm 5cm .5cm 5cm, clip, width=\fs\linewidth]{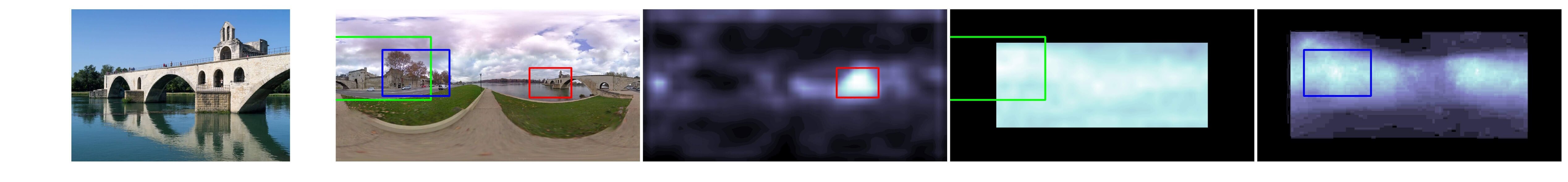}{(l)}\\
    \includegraphics[trim=.5cm 5cm .5cm 5cm, clip, width=\fs\linewidth]{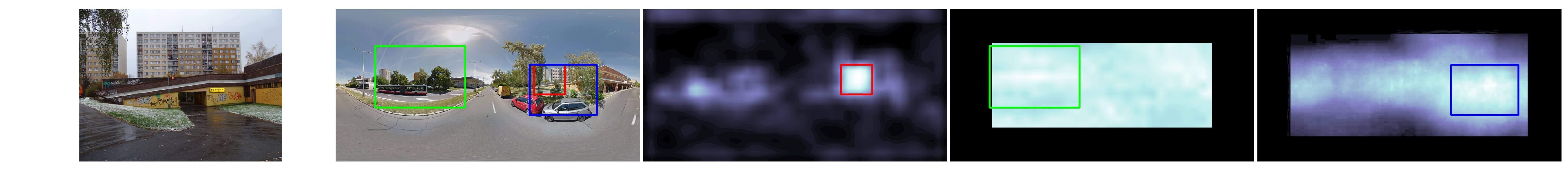}{(m)}\\
    \includegraphics[trim=.5cm 5cm .5cm 5cm, clip, width=\fs\linewidth]{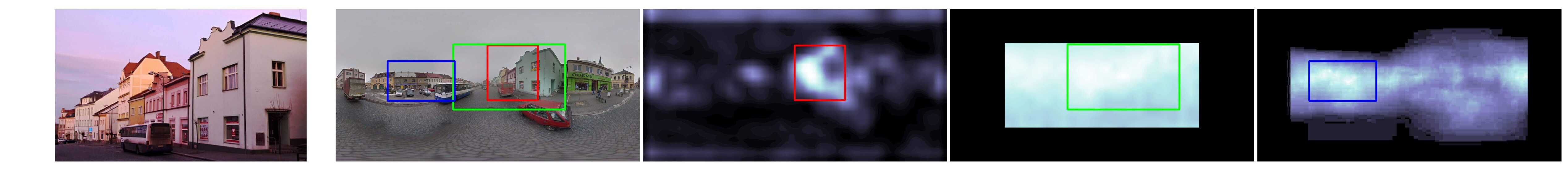}{(n)}\\   
    \includegraphics[trim=.5cm 5cm .5cm 5cm, clip, width=\fs\linewidth]{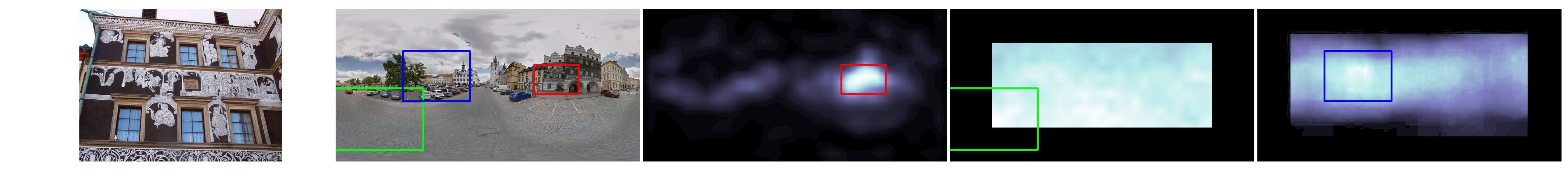}{(o)}\\ 
    \includegraphics[trim=.5cm 5cm .5cm 5cm, clip, width=\fs\linewidth]{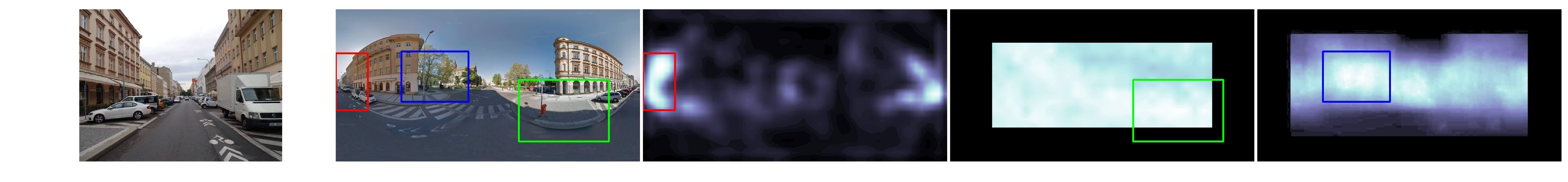}{(p)}\\
    \caption{Localization performance comparisons for BUPM, BBS and DDIS. (Best viewed in digital version. Zoom in for details) Bounding box color: \textcolor{red}{\mysquare} BUPM, \textcolor{green}{\mysquare} BBS, \textcolor{blue}{\mysquare} DDIS}
    \label{fig:localization}
\end{figure}

Due to the lack of ground truth annotation on matched regions, we only assess the localization performance qualitatively. Fig.~\ref{fig:localization} shows the localization results of the proposed BUPM network, BBS, and DDIS, as well as corresponding probability maps where a brighter pixel indicates a higher likelihood of matching. The provided bounding boxes of BUPM are obtained by finding the minimum rectangular box of the biggest connected component on the mask (after thresholding the predict mask at 0.5), while those of BBS and DDIS are directly taken from the source code outputs.

It is clear that the proposed BUPM network 1) produces fewer bright pixels and thus fewer false alarms than BBS and DDIS; 2) focuses more on immovable objects like buildings as expected, while BBS and DDIS are not (see  Fig.~\ref{fig:localization}-(c,h)); 3) is more robust against natural variations like day-night change (see Fig.~\ref{fig:localization}-(d,h,j,l,n,p)) weather change (see Fig.~\ref{fig:localization}-(c)), and season change (see Fig.~\ref{fig:localization}-(m)); 4) is capable to handle the extreme case that a panorama image is opened at some place inside of the query scene, which will cause the matched content in a reference locates at both the left- and right-most regions (see Fig.~\ref{fig:localization}-(p)). 

\section{Conclusion}~\label{sec.conclude}
In this paper, we work on the image-to-GPS verification task to fight against fake news. We show that this problem could be formulated as an image verification problem -- whether or not a query image and a reference image retrieved at the claimed position are taken from the same position. More precisely, we answer this question by checking visual contents of a query image in one panorama reference image, instead of verifying the query image against many possible reference images. This novel setting not only reduces the number of verification need to perform but also largely simplifies data preparation. 

We propose a novel BUPM network to perform the verification task. It is a feed-forward network with multiple stages, each of which is designed to fulfill one particular goal, \eg{} visual feature extraction, similarity matching, etc. Since there is no large enough public dataset for training BUPM, we collect and clean 30K paired positive samples. To ensure the BUPM module to achieve its designed functionality and network convergence, we also introduce a two-stage training scheme. Our experimental results on the real dataset demonstrate that the proposed BUPM network outperforms state-of-the-art image verification methods in terms of much higher AUC and average precision scores and that it is capable of finding matched patches between query and reference. The bottom-up matching manner further improves matching accuracy and reduces false alarms. 

Since the BUPM network solution can be viewed as one way of learnable template matching, it can be applied to related problems, like template matching, constraint splicing detection~\cite{wu2017deep}, etc.

\bigskip
\noindent {\bf Acknowledgement} This work is based on research sponsored by the Defense Advanced Research Projects Agency under agreement number FA8750-16-2-0204. The U.S. Government is authorized to reproduce and distribute reprints for governmental purposes notwithstanding any copyright notation thereon. The views and conclusions contained herein are those of the authors and should not be interpreted as necessarily representing the official policies or endorsements, either expressed or implied, of the Defense Advanced Research Projects Agency or the U.S. Government.

\clearpage
\bibliographystyle{splncs04}
\bibliography{egbib}
%




\end{document}